\documentclass[journal]{IEEEtran}

\usepackage{enumitem}
\usepackage{placeins}
\usepackage{afterpage}
\usepackage{float}
\usepackage{booktabs}
\usepackage{cite}
\usepackage{graphicx}
\usepackage[caption=false,font=footnotesize]{subfig}
\usepackage{amsmath}
\usepackage{xcolor} 
\usepackage[hidelinks]{hyperref}
\begin{document}

\title{Absorption-Feature-Guided Distance-Decoupled Estimation and Band Selection for LWIR Hyperspectral Passive Ranging}

\author{
	Shuo~Liu,
	Chen~Fan,
	Zhihe~Chen,
	Xiaolin~Huang,
	and~Lilian~Zhang%
	\thanks{This work was supported in part by the National Natural Science Foundation of China under Grant 62673486, in part by the Innovation Research Foundation of National University of Defense Technology under Grant 24-ZZCX-JDZ-28.}%
	\thanks{Shuo Liu and Chen Fan contributed equally to this work.}%
	\thanks{Shuo Liu, Chen Fan, Zhihe Chen, Xiaolin Huang, and Lilian Zhang are with the College of Intelligence Science and Technology, National University of Defense Technology, Changsha, China. Lilian Zhang is also with the School of Future Technology, National University of Defense Technology, Changsha, China.}%
	\thanks{Corresponding author: Lilian Zhang. E-mail: lilianzhang@nudt.edu.cn.}%
	\thanks{%
		\copyright\ 2026 IEEE. Personal use of this material is permitted.
		Permission from IEEE must be obtained for all other uses,
		in any current or future media, including
		reprinting/republishing this material for advertising
		or promotional purposes, creating new collective works,
		for resale or redistribution to servers or lists,
		or reuse of any copyrighted component of this work
		in other works.
	}%
}

\markboth
{Absorption-Feature-Guided Distance-Decoupled Estimation and Band Selection for LWIR Hyperspectral Passive Ranging}
{Liu \MakeLowercase{\textit{et al.}}: Absorption-Feature-Guided Distance-Decoupled Estimation and Band Selection for LWIR Hyperspectral Passive Ranging}

\IEEEtitleabstractindextext{%
	\begin{abstract}
		In long-wave infrared (LWIR) hyperspectral observations, atmospheric absorption varies with propagation distance and imprints distance-dependent spectral signatures on the observed radiance, providing a physical basis for passive ranging.
		In natural scenes, however, these signatures are nonlinearly coupled with target temperature, material emissivity, and path radiance, making distance inversion ill-posed.
		Existing methods typically rely on full-band measurements and pixel-wise gradient-based joint optimization of distance, temperature, and emissivity, resulting in high computational cost without explicitly exploiting sharp atmospheric absorption structures for efficient ranging.
		Therefore, we propose an Absorption-Guided Distance-Decoupled Estimation and Refinement (ADER) framework.
		ADER incorporates a spectral-smoothness prior by representing material emissivity using B-spline control points, thereby suppressing overfitting to sharp atmospheric absorption structures and facilitating distance-decoupled estimation from the observed radiance.
		Using ozone-absorption cues to identify reflected downwelling radiance, ADER classifies pixels as emission-dominant or reflection-dominant.
		For emission-dominant pixels, ADER compensates for path radiance and transmittance and estimates distance through one-dimensional minimization of the resulting absorption residual; for reflection-dominant pixels, it refines the initial distance estimate using the complete radiative model with downwelling-radiance compensation.
		To reduce spectral redundancy and enable efficient few-band ranging, we further propose a greedy band selection strategy based on multi-scene effective Fisher information for the distance parameter. 
			Real-scene experiments show that, in the representative LiDAR-referenced grass region, the mean distances estimated by full-band and 20-band ADER deviate from the LiDAR reference by \(0.20~\mathrm{m}\) and \(0.40~\mathrm{m}\), respectively.
			Compared with the reproduced hyperspectral baseline, full-band and 20-band ADER achieve approximately \(130\times\) and \(179\times\) speedups, respectively.
		
	\end{abstract}
	
	\begin{IEEEkeywords}
		Atmospheric absorption, distance-decoupled estimation, band selection, LWIR hyperspectral imaging, downwelling radiance.
	\end{IEEEkeywords}
}

\maketitle

\IEEEdisplaynontitleabstractindextext

\IEEEpeerreviewmaketitle


\begin{figure*}[!t]
	\centering
	\setlength{\abovecaptionskip}{4pt}
	\setlength{\belowcaptionskip}{0pt}
	
	\includegraphics[width=0.98\textwidth]{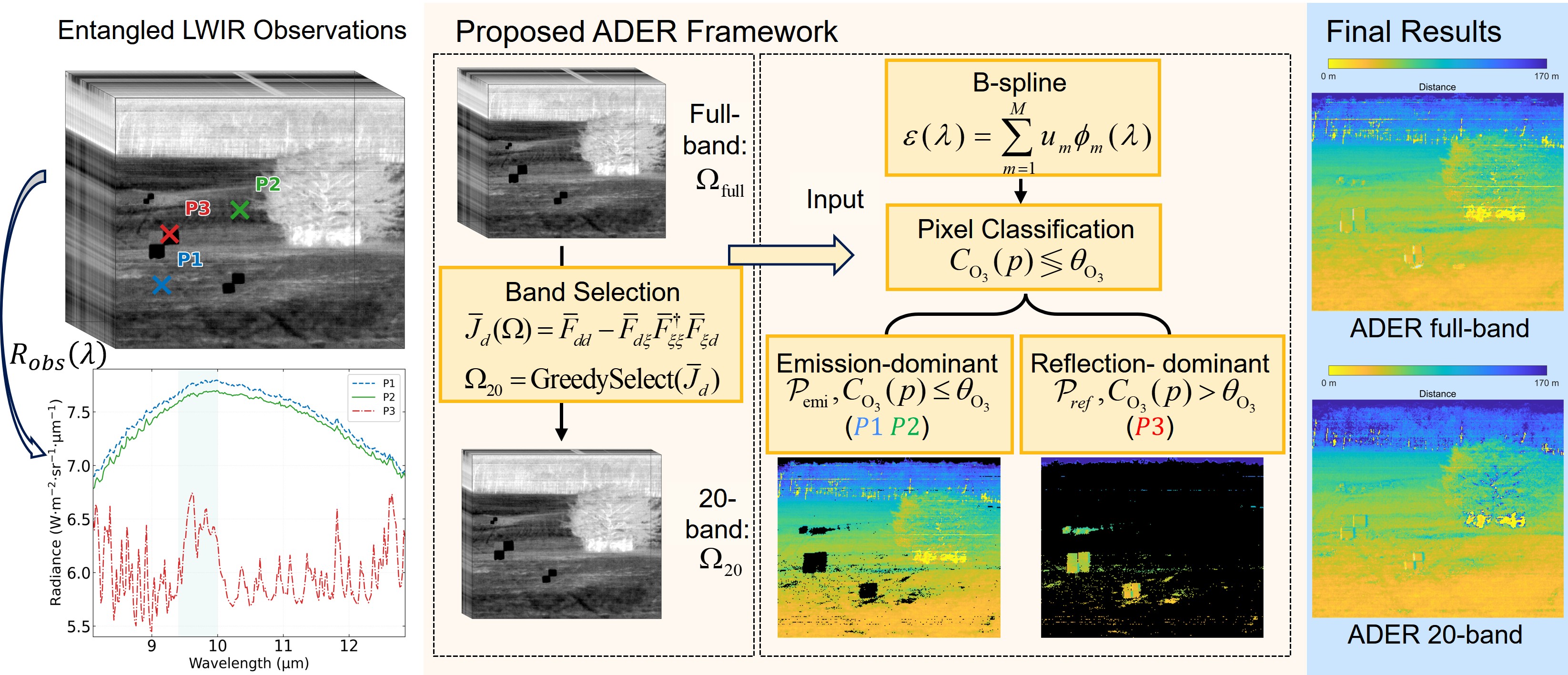}
	
	\vspace{1mm}
	\caption{Motivation, pipeline, and representative ranging results of the proposed ADER framework. 
		The left part shows selected pixels and their corresponding radiance spectra from a real LWIR hyperspectral observation, illustrating the coupling among distance, temperature, and material emissivity in the observed radiance. 
		The middle part summarizes the main ADER pipeline, where either full-band spectra or selected 20-band spectra are used as input for B-spline emissivity modeling, ozone-feature-based pixel classification, and pixel-wise ranging. 
		Emission-dominant pixels, such as P1 and P2, are processed by distance-decoupled estimation, whereas reflection-dominant pixels, such as P3, are processed by distance refinement with downwelling-radiance compensation. 
		The right part presents representative ADER full-band and ADER 20-band ranging results.}
	\label{fig:fig11}
	\vspace{-0.6em}
\end{figure*}

\section{Introduction}
\label{sec:introduction}

\IEEEPARstart{R}{eliable} distance estimation is fundamental to topographic mapping, ground-target observation, and environmental sensing.
Existing ranging methods usually rely on either active sensors or visible-spectrum geometric cues. 
Active sensors, such as LiDAR, provide accurate range measurements but are constrained by power consumption, eye safety, and operational covertness~\cite{royo_overview_2019}. 
Passive geometric methods, including stereo vision and monocular depth estimation, remain vulnerable to low illumination, fog, and texture deficiency~\cite{wang_regularizing_2021,yao_foggystereo_2022}. 
These limitations motivate passive ranging methods that do not rely on active illumination or visible-spectrum texture~\cite{bao_heat-assisted_2023}.

LWIR hyperspectral imaging provides a physics-based route to passive ranging beyond geometry-based vision by exploiting radiative propagation~\cite{nagase_shape_2022,dorken_gallastegi_absorption-based_2025}. 
At ambient temperature, the thermal emission peak of natural targets falls within the LWIR spectral range, enabling passive observation of target radiance under low-light or no-light conditions~\cite{manolakis_longwave_2019}.
Meanwhile, hyperspectral imaging provides fine spectral resolution and can capture the influence of absorption features from gases such as water vapor, carbon dioxide, and ozone on target radiance~\cite{manolakis_hyperspectral_2016,eismann_hyperspectral_2012}. 
	Recent hyperspectral and multispectral perception studies have increasingly exploited spectral complementarity and spectral--spatial information to improve visual tracking and object detection under complex conditions~\cite{zhao_deptharea_2026,zhao_sasunet_2025,zhang_cross_domain_hsi_detection_2026}.
For passive ranging, however, spectral information is exploited through a different physical mechanism: according to the Beer--Lambert law, atmospheric absorption varies with propagation distance and leaves distance-dependent atmospheric absorption signatures in the observed spectrum, which provides the physical basis for LWIR hyperspectral passive ranging~\cite{dorken_gallastegi_absorption-based_2024}.

However, converting these distance-dependent absorption signatures into pixel-wise distance estimates remains challenging. 
The observed radiance combines target emission, atmospheric path radiance, and reflected environmental radiance~\cite{manolakis_hyperspectral_2016,eismann_hyperspectral_2012,young_scene_2002}.
Therefore, target distance \(d\), target temperature \(T\), and wavelength-dependent emissivity \(\varepsilon(\lambda)\) are nonlinearly coupled in the observed spectrum.
Different combinations of these physical variables may produce similar spectral responses, making distance recovery an underdetermined and ill-conditioned inverse problem~\cite{dorken_gallastegi_absorption-based_2024,ren_feature_band_hybrid_2022}.
Recent work has demonstrated that atmospheric absorption signatures in LWIR hyperspectral observations can support passive ranging in natural scenes by jointly estimating distance, temperature, and per-band emissivity from full-band measurements~\cite{dorken_gallastegi_absorption-based_2025}.
However, existing LWIR hyperspectral ranging methods mainly treat atmospheric absorption as one component of a full-band joint inversion model.
Their per-band emissivity representation introduces excessive degrees of freedom, making it difficult to use the contrast between smooth material emissivity and sharp atmospheric absorption structures to isolate distance information.
The resulting high-dimensional pixel-wise optimization also limits computational efficiency.
Moreover, pixels in natural scenes may have different dominant radiative mechanisms. 
For most high-emissivity targets, the reflected radiance can be neglected and the observed radiance can be simplified to target emission and path radiance~\cite{dorken_gallastegi_absorption-based_2025}. 
In contrast, for reflection-dominant regions such as reflective boards and sky boundaries, neglecting reflected downwelling radiance can cause distance overestimation~\cite{gallastegi_ozone_2026}. 
Using a unified model for all pixels therefore makes it difficult to balance model completeness and computational efficiency. 
This motivates pixel classification according to different dominant radiative mechanisms and the design of corresponding distance estimation strategies.
In addition, although full-band hyperspectral observations provide richer spectral information, they also introduce substantial spectral redundancy and unnecessary data acquisition and computational burden~\cite{bioucas-dias_hyperspectral_2013,sun_hyperspectral_2019,zhao_tir_band_selection_2024}.
Most existing hyperspectral band selection methods target classification, target detection, or spectral unmixing rather than the observability of a physical distance parameter.
At the other extreme, bispectral and quadspectral ranging are computationally efficient but may sacrifice ranging accuracy in complex real scenes~\cite{gallastegi_ozone_2026}.
Therefore, LWIR hyperspectral passive ranging requires a task-oriented band selection strategy that reduces spectral redundancy while preserving distance observability.

These limitations indicate that LWIR hyperspectral passive ranging should not rely on full-band numerical optimization alone.
Instead, distance information should be isolated by explicitly exploiting the structural difference between sharp atmospheric absorption and smooth material emissivity. To this end, we propose an Absorption-Guided Distance-Decoupled Estimation and Refinement (ADER) framework for LWIR hyperspectral passive ranging.
ADER incorporates a spectral-smoothness prior by representing material emissivity using B-spline control points.
This low-dimensional representation reduces emissivity unknowns and suppresses the fitting of sharp atmospheric absorption structures by the emissivity estimate.
Following prior work on ozone-feature cues for identifying reflected downwelling radiance~\cite{gallastegi_ozone_2026}, ADER uses this cue to classify scene pixels into emission-dominant and reflection-dominant groups.
For emission-dominant pixels, ADER compensates path radiance and transmittance, extracts the absorption residual, and estimates distance by a one-dimensional residual minimization.
For reflection-dominant pixels, the downwelling-radiance representation introduced in prior work~\cite{gallastegi_ozone_2026} is incorporated into the complete radiative model to refine the distance initialized by the proposed distance-decoupled estimate.
Furthermore, we design a greedy band selection strategy based on multi-scene effective Fisher information for the distance parameter to reduce spectral redundancy.
	Experiments show that ADER recovers spatial distance structures consistent with the approximately registered LiDAR reference under both full-band and 20-band settings, while providing improved region-level distance statistics over the evaluated baseline within the selected quantitative evaluation regions and substantially reducing computational cost.
Fig.~\ref{fig:fig11} gives an overview of the motivation, processing pipeline, and representative results of ADER.

The main contributions of this paper are summarized as follows:
\begin{itemize}[leftmargin=1.2em,itemsep=0.25em,topsep=0.25em]
	\item \textcolor{black}{
		To exploit the structural difference between smooth
		material emissivity and sharp atmospheric absorption
		for distance decoupling, we represent material emissivity
		using B-spline control points.
		This low-dimensional representation reduces the number
		of emissivity unknowns and provides the smooth spectral
		subspace used for absorption-residual projection in the
		proposed distance-decoupled estimator.
	}
	\item To avoid high-dimensional optimization over all image pixels and improve computational efficiency, we propose a pixel-classification-based distance estimation and refinement strategy.
	ADER uses existing downwelling-radiance cues to distinguish emission-dominant and reflection-dominant pixels.
	For emission-dominant pixels, which constitute the majority of scene pixels, we propose absorption-residual-based distance-decoupled estimation, which formulates distance estimation as a one-dimensional search problem, thereby significantly reducing computational complexity; for reflection-dominant pixels, the complete radiative model with downwelling-radiance compensation is further used for refinement, initialized by the distance-decoupled estimate.
	
	\item To reduce spectral redundancy and enable efficient few-band ranging, we propose a greedy band selection strategy based on multi-scene effective Fisher information for the distance parameter. 
	By maximizing the incremental information gain, the proposed strategy selects distance-sensitive bands with complementary information. 
	Experiments show that approximately 20 selected bands maintain ranging results close to the 256-band setting, reducing the number of used bands by about one order of magnitude.
\end{itemize}

The remainder of this paper is organized as follows. 
Section~\ref{sec:related_work} reviews related work. 
Section~\ref{sec:radiation_modeling} introduces the radiative model, B-spline emissivity representation, and pixel classification strategy based on downwelling-radiance cues. 
Section~\ref{sec:decoupled_estimation_refinement} analyzes distance information in atmospheric absorption structures and presents the distance-decoupled estimation and reflection-compensated refinement of ADER. 
Section~\ref{sec:band_selection} describes the band selection strategy for the ranging task. 
Section~\ref{sec:experiments} evaluates ADER on real scenes. 
Finally, Section~\ref{sec:conclusion} concludes this paper.


\section{Related Work}
\label{sec:related_work}

\subsection{Absorption-Based Passive Ranging}
\label{subsec:absorption_based_passive_ranging}

Absorption-based passive ranging exploits wavelength-dependent gas attenuation, whose variation with propagation distance embeds range information in the measured radiance spectrum.

Early studies mainly targeted high-temperature sources, including missile plumes, rocket exhausts, jet engines, and thermal lamps, and used \(\mathrm{CO}_2\) or \(\mathrm{O}_2\) absorption for passive ranging~\cite{leonpacher_passive_1983,hawks_passive_2006}. 
Subsequent few-band systems estimated atmospheric transmittance from an absorption band and a nearby reference band and mapped it to distance using a precomputed transmittance--range relation~\cite{hawks_short-range_2013,yu_passive_2017}.

However, this high-temperature-target assumption does not readily extend to natural scenes. 
Targets such as grass and rocks may have temperatures close to the surrounding air, making the distance-dependent absorption modulation weak and difficult to estimate. 
Recent advances in LWIR multispectral and hyperspectral imaging have extended absorption-based ranging to low-temperature-contrast scenes. 
Nagase \emph{et al.} demonstrated the feasibility of passive ranging in real scenes by modeling the attenuation of LWIR thermal radiation through air~\cite{nagase_shape_2022}.
Building on this line of work, Kushida \emph{et al.} introduced an affine-transform representation for multispectral LWIR depth sensing to reduce calibration cost, while enabling closed-form estimation of object distance and temperature~\cite{kushida_affine_2024}. 
Dorken Gallastegi \emph{et al.} analyzed the performance limits of LWIR hyperspectral absorption-based ranging from theoretical and simulation perspectives~\cite{dorken_gallastegi_absorption-based_2024}. 
They further proposed a systematic LWIR hyperspectral passive ranging method, including bispectral and hyperspectral formulations, and estimated distance in real natural scenes within the \(8\)--\(13.2~\mu\mathrm{m}\) band while jointly estimating temperature and emissivity~\cite{dorken_gallastegi_absorption-based_2025}. 
In the hyperspectral formulation, full-band joint optimization and emissivity smoothness regularization were used to improve inversion stability. 
That work also showed that external atmospheric priors are beneficial for distance estimation, and that more spectral observations help reduce range uncertainty caused by temperature contrast and measurement noise. 
More recently, to address reflected downwelling radiance, Gallastegi \emph{et al.} represented the reflected component using downwelling radiance spectra precomputed at different zenith angles, thereby mitigating range overestimation in reflection-dominant regions rather than simply discarding these pixels~\cite{gallastegi_ozone_2026}.

Overall, absorption-based ranging has progressed from few-band ratio methods for high-temperature targets to multispectral and hyperspectral inversion for ambient-temperature natural scenes, with recent studies further accounting for reflected downwelling radiance. 
Nevertheless, existing hyperspectral methods still rely mainly on full-band joint inversion and do not explicitly exploit absorption structures to simplify distance estimation.

\subsection{Atmospheric Compensation and Temperature--Emissivity Separation}
\label{subsec:atmospheric_compensation_tir_inversion}

LWIR hyperspectral remote sensing has been widely used for material identification and physical-parameter retrieval. 
In conventional thermal infrared remote sensing, atmospheric absorption is treated as a propagation effect to be estimated and corrected so that surface-leaving radiance can be recovered. 
Representative atmospheric compensation methods use in-scene near-blackbody targets, water-vapor absorption features, or low-dimensional correlations among atmospheric components~\cite{young_scene_2002,gu_autonomous_2000,acito_coupled_2019,wang_airborne_2024}. Recent noise-resilient AC--TES frameworks further address spectral noise and model--data mismatch across ground-based and airborne LWIR hyperspectral sensors~\cite{wang_noise_resilient_tes_2026}.
In absorption-based ranging, by contrast, the path-length dependence of atmospheric transmittance is itself the source of distance information; the corresponding absorption structures must therefore be preserved and exploited rather than removed.

After atmospheric compensation, temperature--emissivity separation remains underdetermined and requires additional emissivity priors. 
Existing TES methods reduce this ambiguity using empirical emissivity relationships, iterative spectral-smoothness constraints, downwelling-radiance-based refinement, linear or subspace constraints, and dictionary-based representations~\cite{gillespie_temperature_1998,borel_iterative_1997,ingram_sensitivity_2001,cheng_stepwise_tes_2010,wang_temperature_2011,acito_subspace_tes_2019,ma_dictionary_tes_2023}. 
Recent studies have further refined TES for airborne observations and model assumptions~\cite{zhang_airborne_tes_2025,zhang_revisit_tes_2025,zhu_urban_tes_2025}. Physics-based, feature-band, and learning-based schemes have also extended hyperspectral thermal retrieval to selected-band and joint surface--atmosphere inversion~\cite{islam_physics_tes_2017,ren_feature_band_hybrid_2022,li_prior_knowledge_tes_2026}. More recently, ozone-emission effects have been explicitly incorporated into thermal-infrared hyperspectral TES~\cite{zhang_ozone_emission_tes_2026}.

These studies establish emissivity smoothness and low dimensionality as useful physical priors. 
However, existing hyperspectral ranging methods mainly use emissivity smoothness regularization to stabilize full-band joint inversion and do not explicitly isolate distance-related absorption structures from the smooth target spectrum.
This distinction motivates the use of a low-dimensional emissivity representation in absorption-guided distance decoupling.

\subsection{Hyperspectral Perception and Band Selection}
\label{subsec:hyperspectral_dimensionality_reduction}

	Recent hyperspectral and multispectral perception methods have exploited spectral complementarity, adaptive spectral aggregation, spatial--spectral--temporal modeling, depth-assisted scale adaptation, and cross-domain transfer for visual tracking and object detection~\cite{zhao_deptharea_2026,zhao_sasunet_2025,teng_uav_prompt_2026,jiang_crossmodal_2025,zhang_cross_domain_hsi_detection_2026}. 
	Hyperspectral anomaly-detection methods have further employed collaborative spatial--spectral modeling, sliding dual-window-inspired reconstruction, nonlocal--local self-supervised feature coupling, and frequency-domain processing~\cite{zhao_tbcnet_2026,wang_sliding_dual_window_2024,wang_nonlocal_local_had_2025,yao_dfbsnet_2026}. 
	These studies illustrate task-specific exploitation of spectral--spatial information in hyperspectral perception, whereas ADER focuses on physics-based ranging by exploiting distance-dependent atmospheric absorption structures.

Hyperspectral dimensionality reduction is commonly divided into feature extraction and band selection~\cite{bioucas-dias_hyperspectral_2013,rasti_feature_2020}. 
Unlike feature extraction, band selection preserves the original wavelengths and is therefore more suitable for physics-based retrieval.

Recent unsupervised band-selection methods identify representative and complementary bands using reinforcement learning, graph autoencoders, self-representation, or dual-frequency feature fusion~\cite{feng_mr_selection_2022,zhou_iterative_graph_autoencoder_2023,zhang_joint_graph_selfrep_2024,zeng_dual_frequency_band_selection_2025}.
However, these methods mainly optimize classification or reconstruction performance rather than the observability of a physical distance parameter.

For quantitative retrieval, band subsets can instead be selected according to the sensitivity of the target physical parameter, as demonstrated for land-surface temperature estimation~\cite{zhao_tir_band_selection_2024}. 
Task-oriented band selection has also been investigated on LWIR hyperspectral imagery for off-road traversability estimation, where a small band subset retained performance close to the full spectral set~\cite{yellin_concurrent_band_2024}. 
In absorption-based ranging, the selected subset should preserve effective distance information while accounting for nuisance-parameter coupling and redundancy among bands.

\begin{figure*}[!t]
	\centering
	\includegraphics[width=\textwidth]{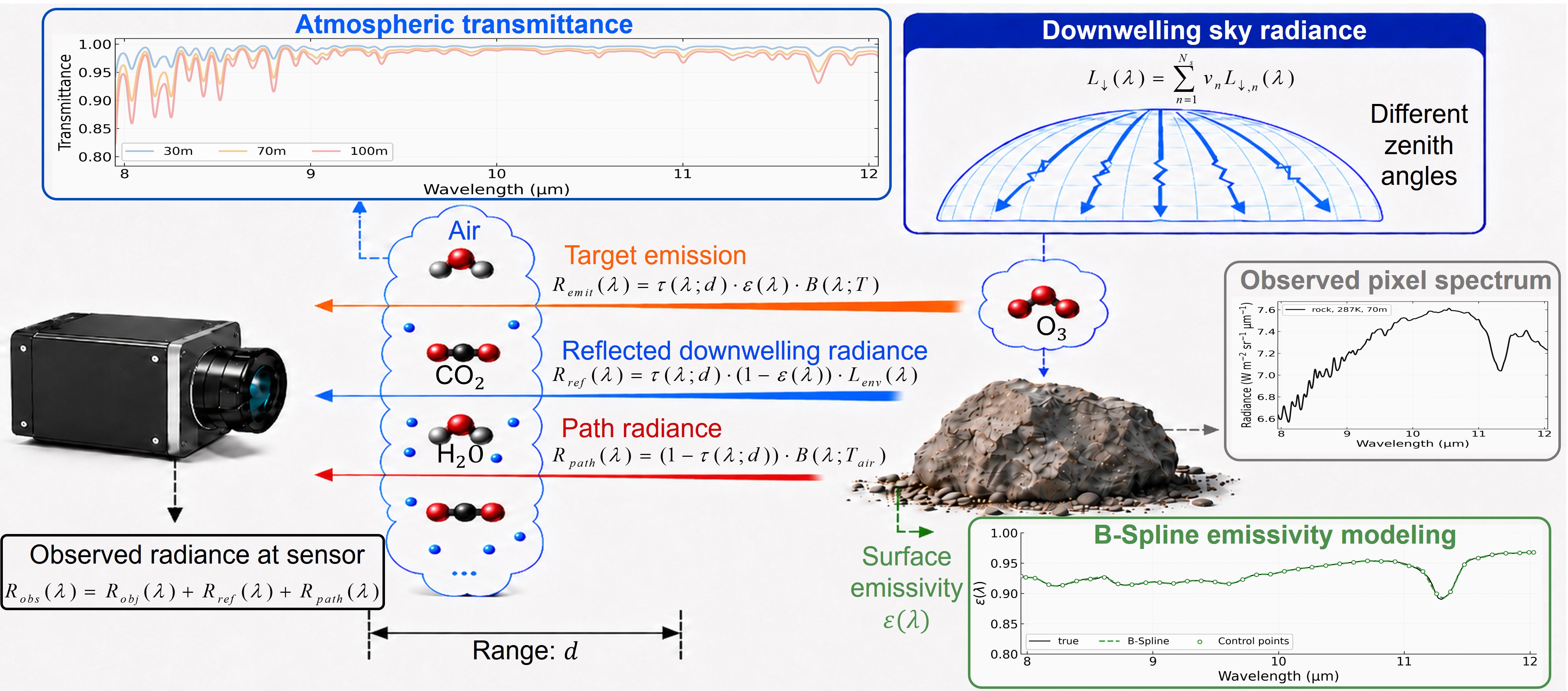}
	\caption{Schematic illustration of the radiative transfer model used in ADER. 
		The observed radiance is decomposed into target emission, reflected downwelling radiance, and path radiance. 
		The sky downwelling radiance is represented by MODTRAN-simulated spectra at different zenith angles~\cite{berk_modtran_2014,gordon_hitran2020_2022}. The ozone absorption feature provides a cue for identifying reflection-dominant pixels with significant reflected downwelling-radiance contribution~\cite{gallastegi_ozone_2026}.
		The surface emissivity is represented using B-spline control points to reduce the number of unknowns and suppress the overfitting of atmospheric absorption structures.}
	\label{fig:fig31}
\end{figure*}

\section{Radiative Model and Pixel Classification}
\label{sec:radiation_modeling}

This section first formulates the radiative transfer model used by ADER. Based on the smoothness prior of material emissivity, the emissivity spectrum is represented using B-splines. 
The ozone absorption feature is further used as a cue for reflected downwelling radiance, enabling pixel classification into emission-dominant and reflection-dominant groups.

\subsection{Standard Radiative Transfer Model}
\label{subsec:standard_radiative_model}

Fig.~\ref{fig:fig31} illustrates the radiative model used in this paper. 
The standard radiative transfer model decomposes the radiance observed by the sensor into three components: target emission, reflected downwelling radiance, and path radiance~\cite{manolakis_hyperspectral_2016,eismann_hyperspectral_2012,young_scene_2002}. 
The first two components originate from the target surface and are modulated by the wavelength-dependent atmospheric transmittance during propagation, while the third component originates from the thermal emission of air along the propagation path. 
In this paper, \(B(\lambda;T)\) denotes the blackbody spectral radiance at wavelength \(\lambda\) and temperature \(T\), and radiance is expressed in \(\mathrm{W\cdot m^{-2}\cdot sr^{-1}\cdot \mu\mathrm{m}^{-1}}\).

When scattering is neglected and the atmosphere is assumed to be approximately homogeneous, the atmospheric transmittance can be described by the Beer--Lambert law. 
We use the unit-path transmittance \(\tau_{1\mathrm{m}}(\lambda)\) to describe distance-dependent attenuation~\cite{nagase_shape_2022,dorken_gallastegi_absorption-based_2024,dorken_gallastegi_absorption-based_2025}, and the transmittance at distance \(d\) is written as
\begin{equation}
	\tau(\lambda;d)=\bigl[\tau_{1\mathrm{m}}(\lambda)\bigr]^d .
	\label{eq:distance_transmittance}
\end{equation}

The complete observed radiance received by the sensor is modeled as
\begin{equation}
	\begin{aligned}
		R_{\mathrm{obs}}(\lambda)
		&=
		R_{\mathrm{emit}}(\lambda)
		+
		R_{\mathrm{ref}}(\lambda)
		+
		R_{\mathrm{path}}(\lambda)\\
		&=
		\tau(\lambda;d)\varepsilon(\lambda)B(\lambda;T)\\
		&\quad+
		\tau(\lambda;d)
		\left(1-\varepsilon(\lambda)\right)
		L_{\downarrow}(\lambda)\\
		&\quad+
		\left(1-\tau(\lambda;d)\right)
		B(\lambda;T_{\mathrm{air}}).
	\end{aligned}
	\label{eq:standard_radiative_model}
\end{equation}
Here, \(R_{\mathrm{emit}}\), \(R_{\mathrm{ref}}\), and \(R_{\mathrm{path}}\) denote the target-emission, reflected-downwelling-radiance, and path-radiance contributions received by the sensor, respectively. 
The variables \(\varepsilon(\lambda)\), \(T\), and \(T_{\mathrm{air}}\) denote the target emissivity, target temperature, and air temperature, respectively. 
In this paper, the reflected radiance is modeled using downwelling radiance spectra at different zenith angles:
\begin{equation}
	L_{\downarrow}(\lambda)
	=
	\sum_{n=1}^{N_s}v_nL_{\downarrow,n}(\lambda),
	\qquad
	v_n\geq0,\quad
	\sum_{n=1}^{N_s}v_n\leq1 .
	\label{eq:downwelling_radiance}
\end{equation}
where \(N_s\) is the number of zenith-angle components in the downwelling-radiance dictionary, \(L_{\downarrow,n}(\lambda)\) denotes the MODTRAN-simulated downwelling radiance spectrum corresponding to the \(n\)-th zenith angle~\cite{berk_modtran_2014}, and \(v_n\) is its weight. 
The weighted combination approximates the effective sky downwelling radiance incident on and reflected by the target surface~\cite{young_scene_2002,cheng_stepwise_tes_2010,gallastegi_ozone_2026}.

In the LWIR band, many natural materials have high emissivity~\cite{meerdink_ecostress_2019,baldridge_aster_2009}. 
For emission-dominant pixels, the reflected radiance is weak and can be approximately neglected. 
The model is then simplified to target emission and path radiance~\cite{dorken_gallastegi_absorption-based_2025}:
\begin{equation}
	\begin{aligned}
		R_{\mathrm{obs}}(\lambda)
		&=
		\tau(\lambda;d)\varepsilon(\lambda)B(\lambda;T)
		+
		\left(1-\tau(\lambda;d)\right)
		B(\lambda;T_{\mathrm{air}}).
	\end{aligned}
	\label{eq:high_emissivity_model}
\end{equation}

The atmospheric state of a scene can be specified using meteorological records such as air temperature, pressure, and humidity. 
Given these parameters, the unit-path atmospheric transmittance can be simulated and used as prior information for distance estimation. 
However, for a single emission-dominant pixel under the simplified model, \(K\) spectral observations still require estimating distance, temperature, and \(K\) per-band emissivity values, resulting in \(K+2\) unknowns. 
The problem is therefore underdetermined and the variables are coupled. 
Existing absorption-based ranging methods usually impose spectral smoothness constraints on emissivity and jointly estimate temperature, distance, and emissivity~\cite{dorken_gallastegi_absorption-based_2025}. 
This motivates an improved emissivity representation that reduces the number of unknowns at the parameter level.

\subsection{B-Spline Emissivity Representation}
\label{subsec:bspline_emissivity}

Emissivity spectra of natural materials in the thermal infrared band are usually smooth and continuous, with limited overall amplitude variation~\cite{meerdink_ecostress_2019,baldridge_aster_2009}. By contrast, atmospheric transmittance is affected by gases such as water vapor, carbon dioxide, and ozone, and exhibits pronounced absorption features at specific wavelengths, leading to sharper spectral structures~\cite{berk_modtran_2014,gordon_hitran2020_2022}. 
This contrast makes a smoothness prior on emissivity physically reasonable.

Instead of estimating emissivity independently at each band and then imposing spectral-smoothness regularization or low-dimensional constraints~\cite{borel_iterative_1997,wang_temperature_2011,acito_subspace_tes_2019,ma_dictionary_tes_2023}, we represent the emissivity spectrum using B-spline control points to better exploit these priors~\cite{xu_universal_2026}.
B-splines have local support and continuous smoothness, and can represent the overall spectral curve with a small number of control points. 
Specifically, for the \(K\) operating bands of the sensor, the emissivity is represented as a linear combination of \(M\) B-spline basis functions:
\begin{equation}
	\varepsilon(\lambda)
	=
	\sum_{m=1}^{M}u_m\phi_m(\lambda),
	\label{eq:bspline_emissivity}
\end{equation}
where \(\phi_m(\lambda)\) denotes the \(m\)-th B-spline basis function, \(u_m\) is the corresponding control coefficient, and \(M\) is the number of control points. 
In general, \(M\ll K\). 
Thus, the original \(K\)-dimensional per-band emissivity parameters are compressed into \(M\)-dimensional control coefficients. 
The B-spline basis functions \(\phi_m(\lambda)\) are precomputed according to the operating wavelength range and sampled bands of the sensor. 

\begin{figure}[!t]
	\centering
	\subfloat[Emissivity fitting]{
		\includegraphics[width=0.95\columnwidth]{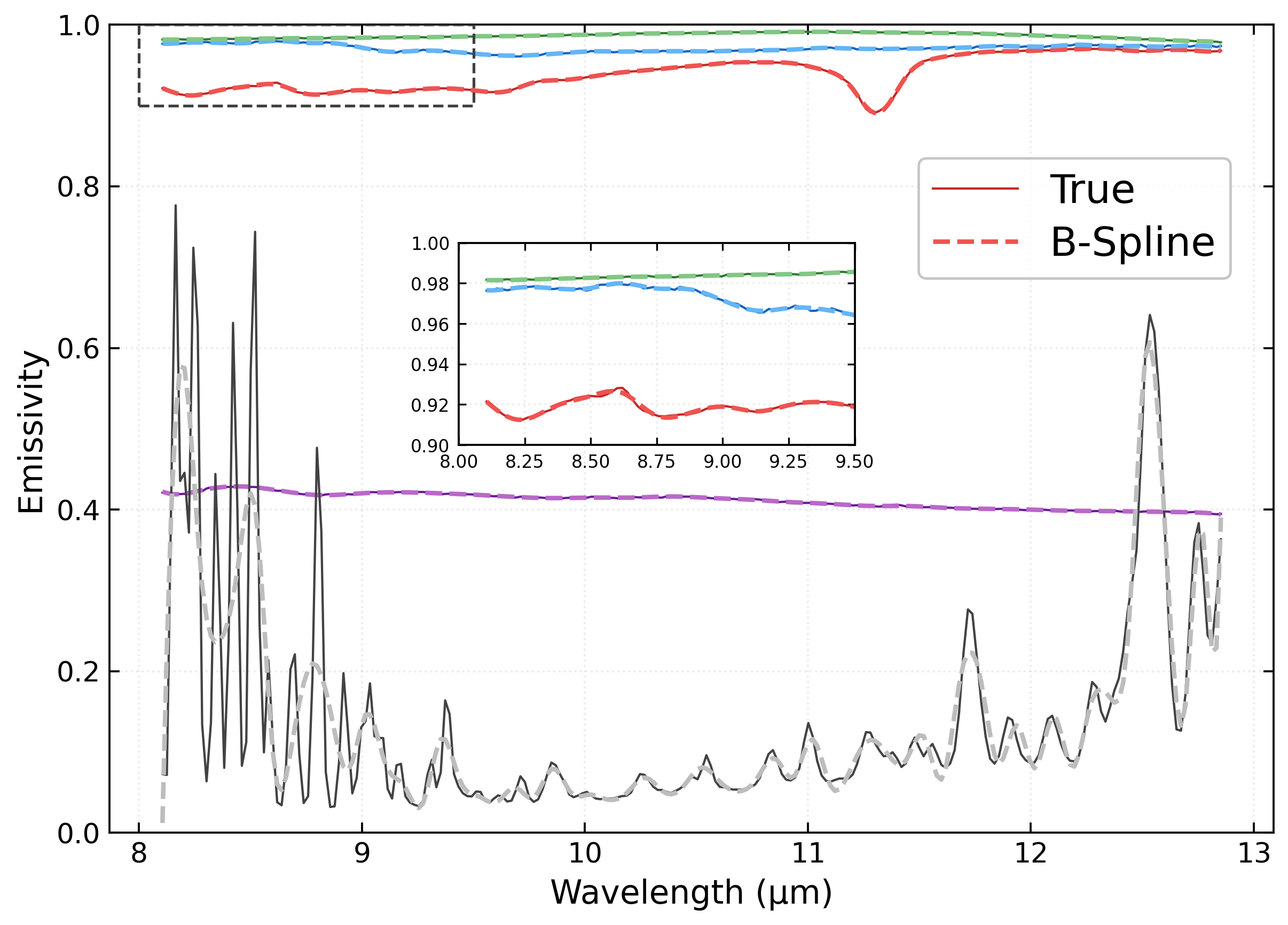}
		\label{fig:fig32a}
	}\\[2pt]
	\subfloat[Fitting error]{
		\includegraphics[width=0.95\columnwidth]{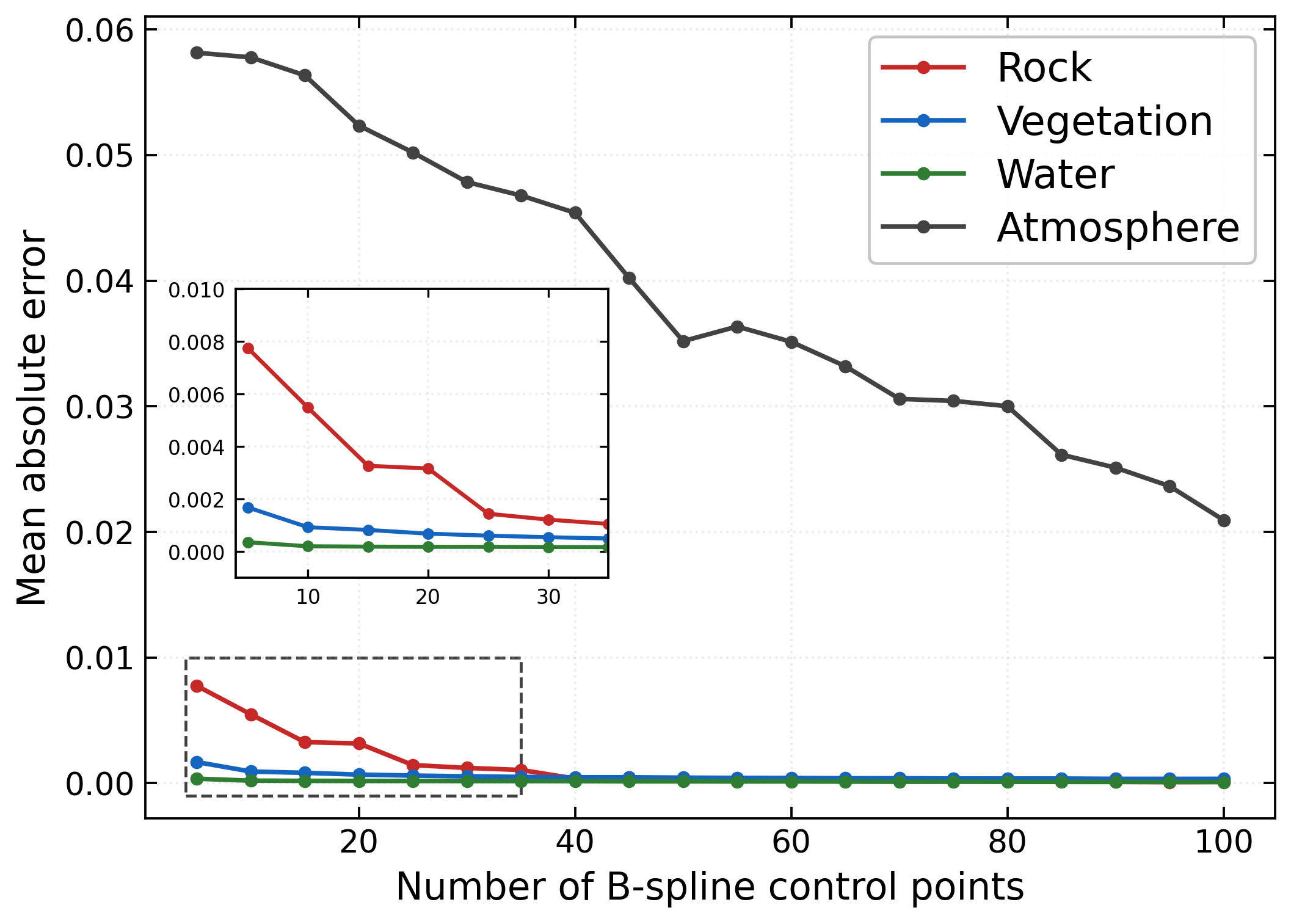}
		\label{fig:fig32b}
	}
	\caption{Comparison of B-spline fitting for material emissivity and atmospheric absorption structures. 
		(a) Representative material emissivity spectra from the ECOSTRESS spectral library~\cite{meerdink_ecostress_2019,baldridge_aster_2009} and their fitting results, together with the atmospheric absorption structure simulated by MODTRAN~\cite{berk_modtran_2014} for comparison. 
		(b) Mean fitting error under different numbers of control points. 
		All high-resolution spectra are first integrated using the sensor instrumental spectral response function with a bandwidth of approximately 40~nm, and then resampled to discrete bands.}
	\label{fig:fig32}
\end{figure}

During inversion, the basis functions are fixed, and only the control coefficients are estimated.
Fig.~\ref{fig:fig32} compares the B-spline fitting results for material emissivity and atmospheric absorption structures. 
To make the spectra consistent with the actual sensor observations, the high-resolution material emissivity spectra from the ECOSTRESS spectral library~\cite{meerdink_ecostress_2019,baldridge_aster_2009} and the high-resolution atmospheric transmittance simulated by MODTRAN~\cite{berk_modtran_2014} are first integrated using the sensor instrumental spectral response function with a bandwidth of approximately 40~nm, and then resampled to 256 discrete bands. 
As shown in Fig.~\ref{fig:fig32}(b), for material emissivity spectra such as rock, vegetation, water, and manmade metal, a small number of control points is sufficient to keep the fitting error below 0.005.
By contrast, under the same number of control points, the fitting error of atmospheric absorption structures is more than one order of magnitude larger.
This indicates that B-splines are suitable for describing smooth material emissivity but have difficulty fitting sharp atmospheric absorption structures, which is beneficial for the subsequent distance-decoupled estimation.

	In this work, the number of B-spline control points is set to \(M=8\) by default.
	The choice balances emissivity representation capability against preservation of distance-dependent atmospheric structure: too few control points may underfit material emissivity, whereas an overly flexible emissivity model can absorb part of the atmospheric structure used for ranging.
	The sensitivity analysis in Appendix~E shows that the emission-dominant regions are relatively stable over the tested values \(M\in\{4,6,8,10,12,16,20\}\), while the reflection-dominant region is more sensitive at the extreme settings.

\subsection{Downwelling-Radiance-Based Pixel Classification}
\label{subsec:pixel_classification}

Real scenes may contain regions strongly affected by reflected downwelling radiance, such as reflective boards and sky boundaries. 
For these reflection-dominant pixels, directly applying the simplified emission-dominant model can lead to biased distance estimation. 
Existing studies have shown that the ozone absorption feature near \(9.6~\mu\mathrm{m}\) in the LWIR band can be used as a cue for identifying reflected downwelling radiance~\cite{gallastegi_ozone_2026}. 
Since ozone is mainly distributed in the upper atmosphere and its absorption contribution along near-ground horizontal paths is weak, an ozone feature observed in the target spectrum is more likely to originate from sky downwelling radiance reflected by the target surface. 
This ozone feature therefore provides a physically grounded cue for identifying pixels strongly affected by reflected downwelling radiance.

\textcolor{black}{
	Unlike existing methods that apply a unified model to the entire image, ADER uses ozone-feature-based pixel classification to assign different distance estimation strategies to emission-dominant and reflection-dominant pixels.
	Following the ozone-feature-based detection approach in prior work~\cite{gallastegi_ozone_2026}, we select two adjacent sampled bands near the ozone absorption feature and compute their local spectral difference as the ozone-feature score.
}
For pixel \(p\), the score is defined as
\begin{equation}
	C_{\mathrm{O}_3}(p)
	=
	\left|
	y_p(\lambda_a)-y_p(\lambda_b)
	\right|,
	\label{eq:ozone_feature_score}
\end{equation}
where \(y_p(\lambda)\) denotes the observed radiance of pixel \(p\) at wavelength \(\lambda\). 
In the current implementation, \(\lambda_a\) and \(\lambda_b\) are the two sampled bands closest to \(9.52~\mu\mathrm{m}\) and \(9.57~\mu\mathrm{m}\), respectively. 
The score is used to characterize the ozone-feature strength caused by reflected downwelling radiance.

The threshold is further normalized using a robust scene-level ozone-feature reference:
\begin{equation}
	\theta_{\mathrm{O}_3}
	=
	\theta_0 C_{\mathrm{O}_3}^{(q)},
	\label{eq:ozone_threshold}
\end{equation}
where \(C_{\mathrm{O}_3}^{(q)}\) denotes the percentile of the image-wide ozone-feature scores at percentile level \(q\). In this work, \(q=99\%\) and \(\theta_0=0.2756\).
According to the comparison between the ozone-feature score and the normalized threshold, the pixel class is determined by
\begin{equation}
	p\in
	\begin{cases}
		\mathcal{P}_{\mathrm{ref}}, & C_{\mathrm{O}_3}(p)>\theta_{\mathrm{O}_3},\\
		\mathcal{P}_{\mathrm{emi}}, & \mathrm{otherwise},
	\end{cases}
	\label{eq:ozone_pixel_classification}
\end{equation}
where \(\mathcal{P}_{\mathrm{ref}}\) denotes the reflection-dominant pixel set and \(\mathcal{P}_{\mathrm{emi}}\) denotes the emission-dominant pixel set.

\begin{figure}[!t]
	\centering
	\includegraphics[width=0.95\columnwidth]{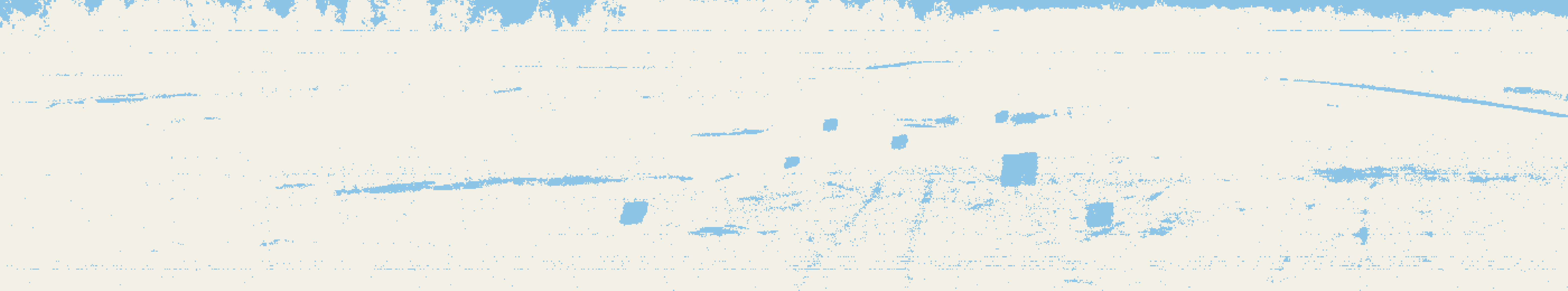}
	\caption{Pixel classification result based on downwelling-radiance cues. 
		The blue regions denote reflection-dominant pixels, while the remaining regions denote emission-dominant pixels.}
	\label{fig:fig33}
\end{figure}

Fig.~\ref{fig:fig33} shows the pixel classification result, where reflection-dominant pixels are marked in blue and the remaining pixels are treated as emission-dominant pixels. 
For emission-dominant pixels, the reflected downwelling radiance is weak, and the simplified radiative model without the reflected term is used by the low-dimensional distance-decoupled estimator. 
For reflection-dominant pixels, the complete radiative model with the reflected term is used for distance refinement. 
The band pair of \(9.52~\mu\mathrm{m}\) and \(9.57~\mu\mathrm{m}\) is not the only possible choice for pixel classification.
The supplementary robustness analysis shows that moderate threshold perturbations mainly alter the classification of isolated pixels, while adjacent-band variations primarily influence local weak-reflection regions; the major reflection-dominant areas remain consistently identified under the tested settings.
In addition, robustness to bright outliers, defective pixels, and spatially nonuniform radiance is evaluated, showing that the percentile-based normalization reduces the sensitivity of the classification threshold to isolated extreme responses while maintaining high overall classification consistency under the tested conditions.
\textcolor{black}{
	The origin and calibration of \(\theta_0\),
	together with the detailed robustness analysis,
	are provided in Appendix~E of the
	supplementary material.
}

\section{Distance-Decoupled Estimation and Reflection-Compensated Refinement}
\label{sec:decoupled_estimation_refinement}

We next analyze how atmospheric absorption structures encode distance information in the observed radiance. 
For emission-dominant pixels, we propose a decoupled distance estimation method based on absorption-residual minimization. 
For reflection-dominant pixels, we introduce reflection-compensated refinement to improve distance estimation under reflected downwelling radiance.

\subsection{Basis for Distance Decoupling}
\label{subsec:physical_basis_distance}

In the emission-dominant radiative model in Eq.~\eqref{eq:high_emissivity_model}, distance information does not appear as an independent additive term in the observed radiance. 
Instead, it modulates the observed radiance through the atmospheric transmittance \(\tau(\lambda;d)\). 
From the perspective of spectral structure, the B-spline emissivity is constrained in a low-dimensional smooth space, and the Planck function also varies smoothly with wavelength. 
Therefore, temperature and emissivity mainly determine the overall radiance magnitude and the slowly varying spectral background. 
By contrast, distance information is related to nonuniform atmospheric absorption and is more strongly reflected in local spectral variations within strong absorption bands. 
Thus, the observed radiance can be approximately interpreted as the combined effect of a smooth background and absorption-related structures, which provides the physical basis for distance decoupling.

\begin{figure}[!t]
	\centering
	\setlength{\abovecaptionskip}{2pt}
	\setlength{\belowcaptionskip}{0pt}
	
	\subfloat[Distance variation]{
		\includegraphics[width=0.95\columnwidth]{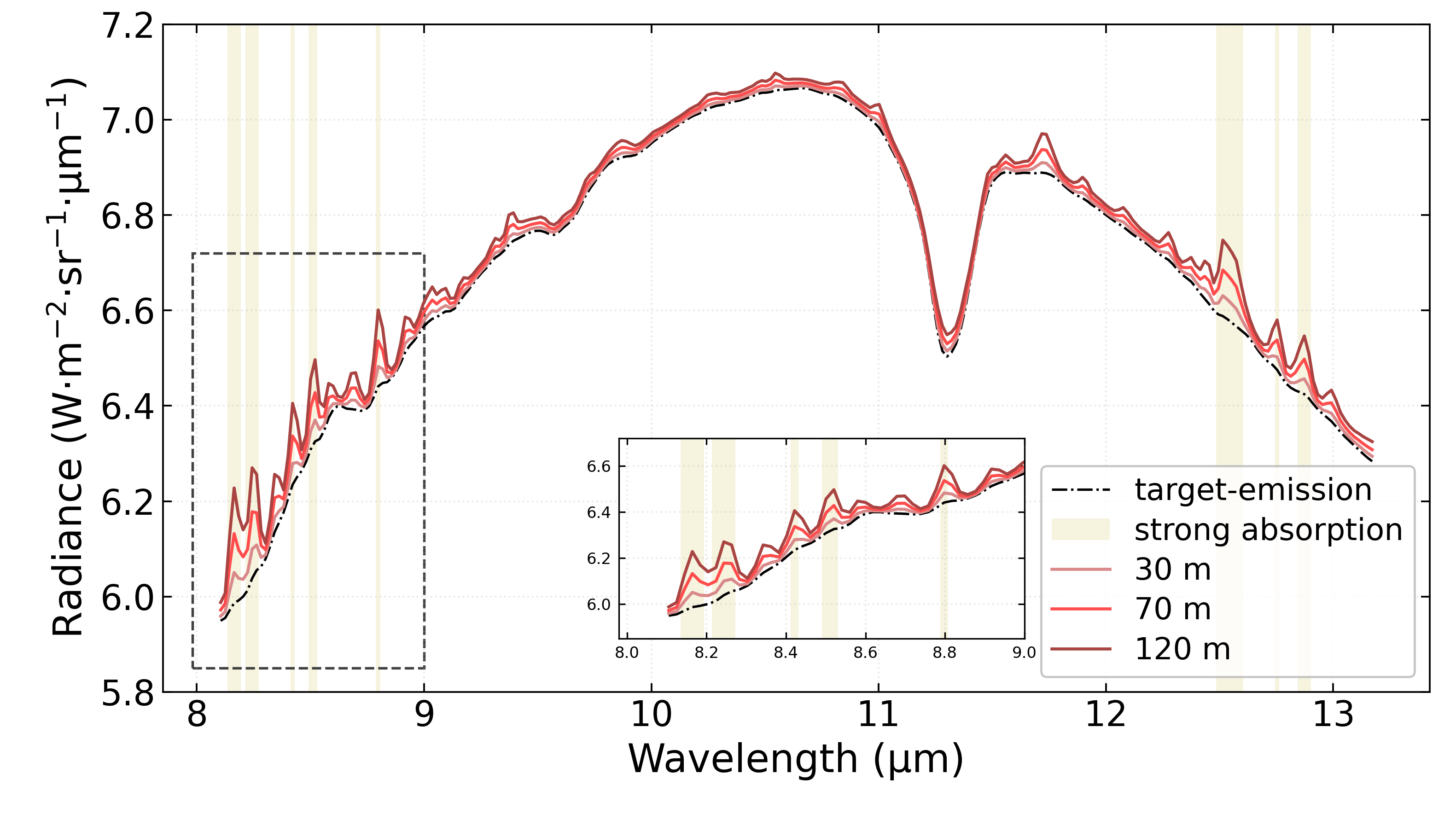}
		\label{fig:fig41a}
	}\\[2pt]
	\subfloat[Temperature variation]{
		\includegraphics[width=0.95\columnwidth]{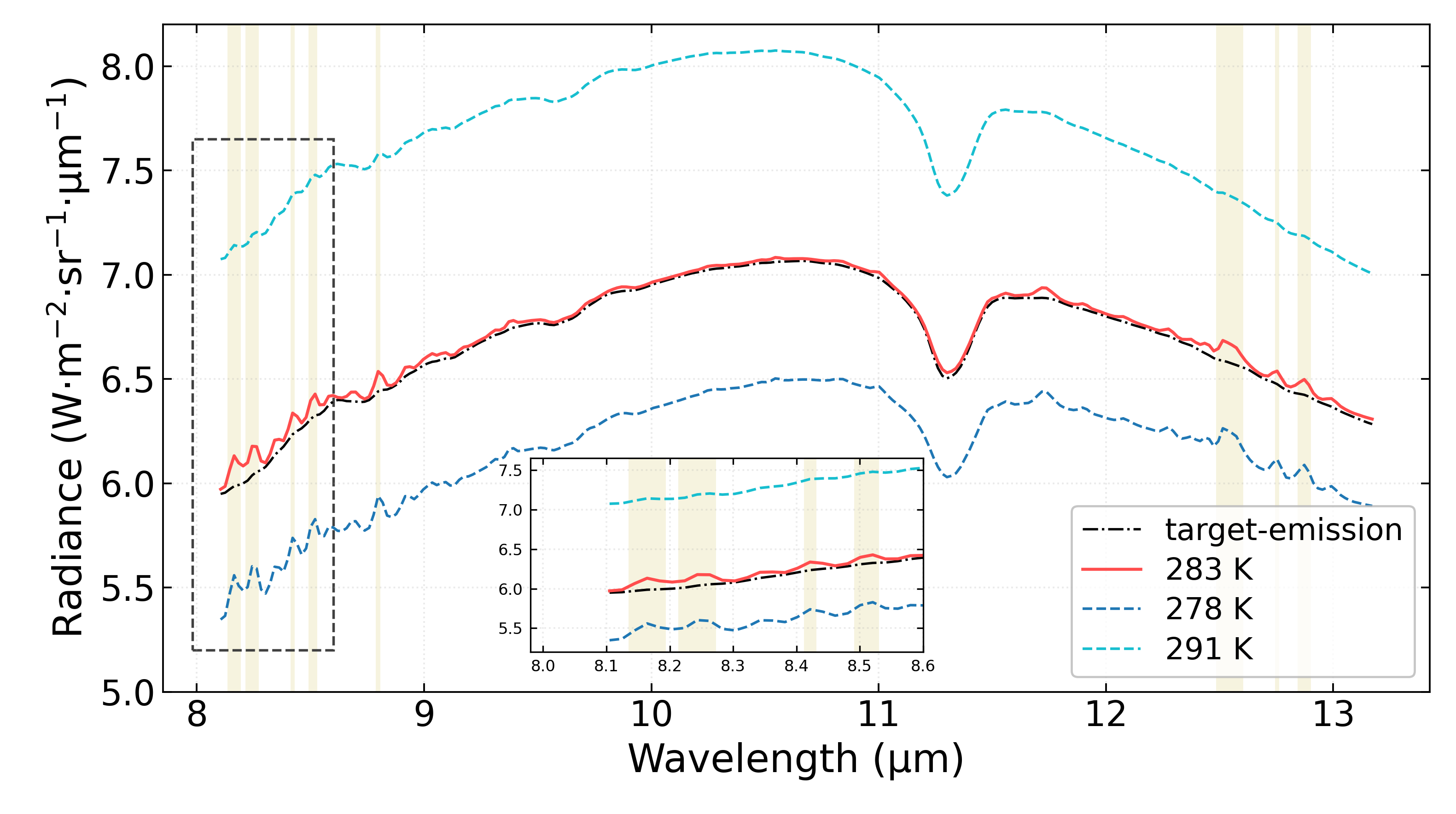}
		\label{fig:fig41b}
	}
	
	\vspace{-0.2em}
	\caption{Radiative simulation analysis using limestone emissivity from the ECOSTRESS spectral library~\cite{meerdink_ecostress_2019,baldridge_aster_2009} and MODTRAN-simulated atmospheric transmittance under the atmospheric setting described in the text~\cite{berk_modtran_2014}. 
		Strong absorption bands are highlighted by shaded background regions. 
		(a) Simulated radiance curves at a target temperature of 283~K and target distances of 30~m, 70~m, and 120~m. 
		The absorption structures become more pronounced as the distance increases. 
		(b) Simulated radiance curves at a fixed distance of 70~m and target temperatures of 278~K, 283~K, and 291~K. 
		Temperature mainly affects the overall radiance magnitude, while a smaller target--air temperature contrast weakens the absorption structures and makes distance recovery more difficult.}
	\label{fig:fig41}
	\vspace{-0.6em}
\end{figure}

Fig.~\ref{fig:fig41} illustrates the different effects of target temperature \(T\), emissivity \(\varepsilon(\lambda)\), and distance \(d\) on the observed radiance. 
Using limestone emissivity from the ECOSTRESS spectral library, we simulate radiance spectra with the radiative model in Eq.~\eqref{eq:high_emissivity_model}. In these simulations, the atmospheric transmittance is generated by MODTRAN with an air temperature of \(290~\mathrm{K}\), a pressure of \(1010~\mathrm{mb}\), a water-vapor partial pressure of \(12.12~\mathrm{mb}\), and a reference path length of \(100~\mathrm{m}\)~\cite{berk_modtran_2014}. 
The high-resolution transmittance is convolved with a Gaussian instrumental spectral response of approximately \(40~\mathrm{nm}\) and then resampled to the sensor bands.
When the target temperature is fixed and the distance increases, the absorption structures in the simulated radiance become more pronounced, especially within the strong absorption bands. 
When the target distance is fixed and the temperature changes, the radiance mainly varies in overall magnitude. 
Moreover, a larger target--air temperature difference leads to more visible distance-related absorption structures, whereas a smaller temperature contrast weakens these structures and makes distance recovery more difficult.

Although the above analysis mainly applies to emission-dominant pixels, it reveals how distance information differs from temperature and emissivity in the observed radiance. 
This difference motivates the distance-decoupled estimator introduced in the following subsection.

\subsection{Distance-Decoupled Estimation}
\label{subsec:absorption_guided_decoupled_estimation}

For a given pixel, \(y_{\mathrm{obs}}(\lambda)\) denotes the measured radiance corresponding to \(R_{\mathrm{obs}}(\lambda)\), and \(y_k=y_{\mathrm{obs}}(\lambda_k)\) denotes its value at the \(k\)-th sampled band. For emission-dominant pixels, we rearrange Eq.~\eqref{eq:high_emissivity_model} under a candidate distance \(d\) to compensate for path radiance and atmospheric attenuation:
\begin{equation}
	\widetilde{y}(\lambda;d)
	=
	\frac{
		y_{\mathrm{obs}}(\lambda)
		-
		\bigl(1-\tau(\lambda;d)\bigr)B(\lambda;T_{\mathrm{air}})
	}{
		\tau(\lambda;d)
	}
	\approx
	\varepsilon(\lambda)B(\lambda;T).
	\label{eq:compensated_radiance}
\end{equation}
When the candidate distance is close to the true distance, the path-radiance term and the propagation attenuation are properly compensated. 
The resulting \(\widetilde{y}(\lambda;d)\) is then mainly determined by target self-emission and becomes closer to a smooth background governed by temperature and emissivity. 
This operation is similar in form to atmospheric compensation. 
However, conventional atmospheric compensation aims to suppress atmospheric absorption effects, whereas the proposed method deliberately exploits the residual absorption structures as distance cues and searches for the distance that makes the compensated spectrum smoothest~\cite{young_scene_2002,gu_autonomous_2000,acito_coupled_2019}.

To further extract the distance-related component from the compensated radiance, this paper defines the residual projection operator as
\begin{equation}
	P_{\mathrm{res}}
	=
	I-\Phi\Phi^{\dagger},
	\label{eq:projection_res}
\end{equation}
where \(\Phi\) is the B-spline basis matrix sampled at the discrete sensor wavelengths, with \([\Phi]_{k,m}=\phi_m(\lambda_k)\), and \(\Phi^{\dagger}\) denotes its Moore--Penrose pseudoinverse. 

	The operator \(\Phi\Phi^{\dagger}\) projects the compensated spectrum onto the B-spline subspace spanned by \(\Phi\), whereas \(P_{\mathrm{res}}=I-\Phi\Phi^{\dagger}\) retains the component outside this smooth spectral subspace.
	This residual therefore contains spectral structures that cannot be represented by the B-spline model, including uncompensated atmospheric absorption features caused by an incorrect candidate distance.
Let \(\widetilde{\mathbf{y}}(d)=
[\widetilde{y}(\lambda_1;d),\cdots,\widetilde{y}(\lambda_K;d)]^{\top}\) denote the compensated spectral vector.
The compensated spectrum can then be decomposed as
\begin{equation}
	\widetilde{\mathbf{y}}(d)
	=
	\underbrace{
		P_{\mathrm{res}}\widetilde{\mathbf{y}}(d)
	}_{\text{absorption residual component}}
	+
	\underbrace{
		\bigl(I-P_{\mathrm{res}}\bigr)\widetilde{\mathbf{y}}(d)
	}_{\text{smooth background component}} .
	\label{eq:compensated_decomposition}
\end{equation}

\begin{figure}[!t]
	\centering
	\setlength{\abovecaptionskip}{2pt}
	\setlength{\belowcaptionskip}{0pt}
	
	\subfloat[Compensated spectra]{
		\includegraphics[width=0.95\columnwidth]{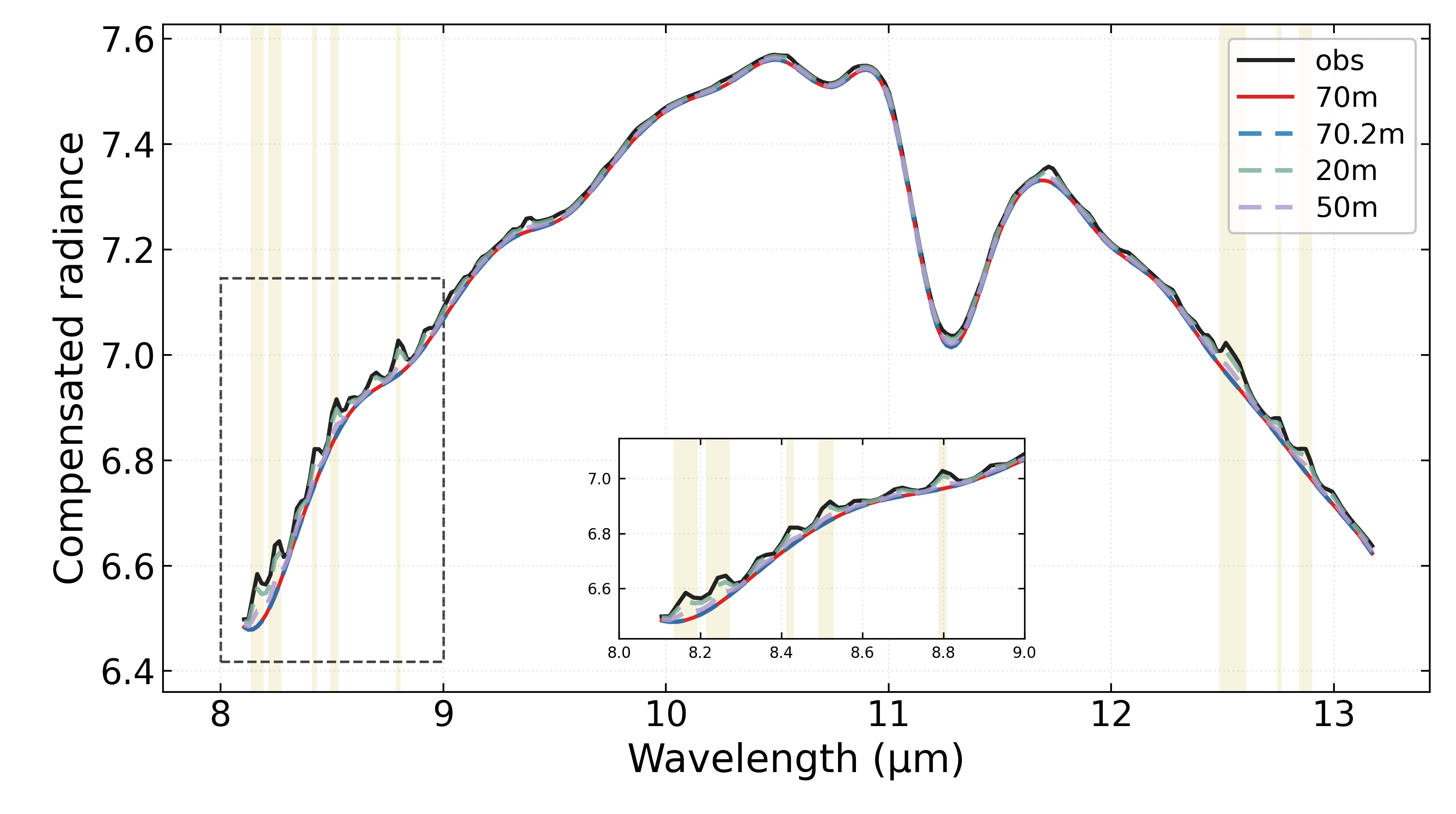}
		\label{fig:fig43a}
	}\\[2pt]
	\subfloat[Absorption residuals]{
		\includegraphics[width=0.95\columnwidth]{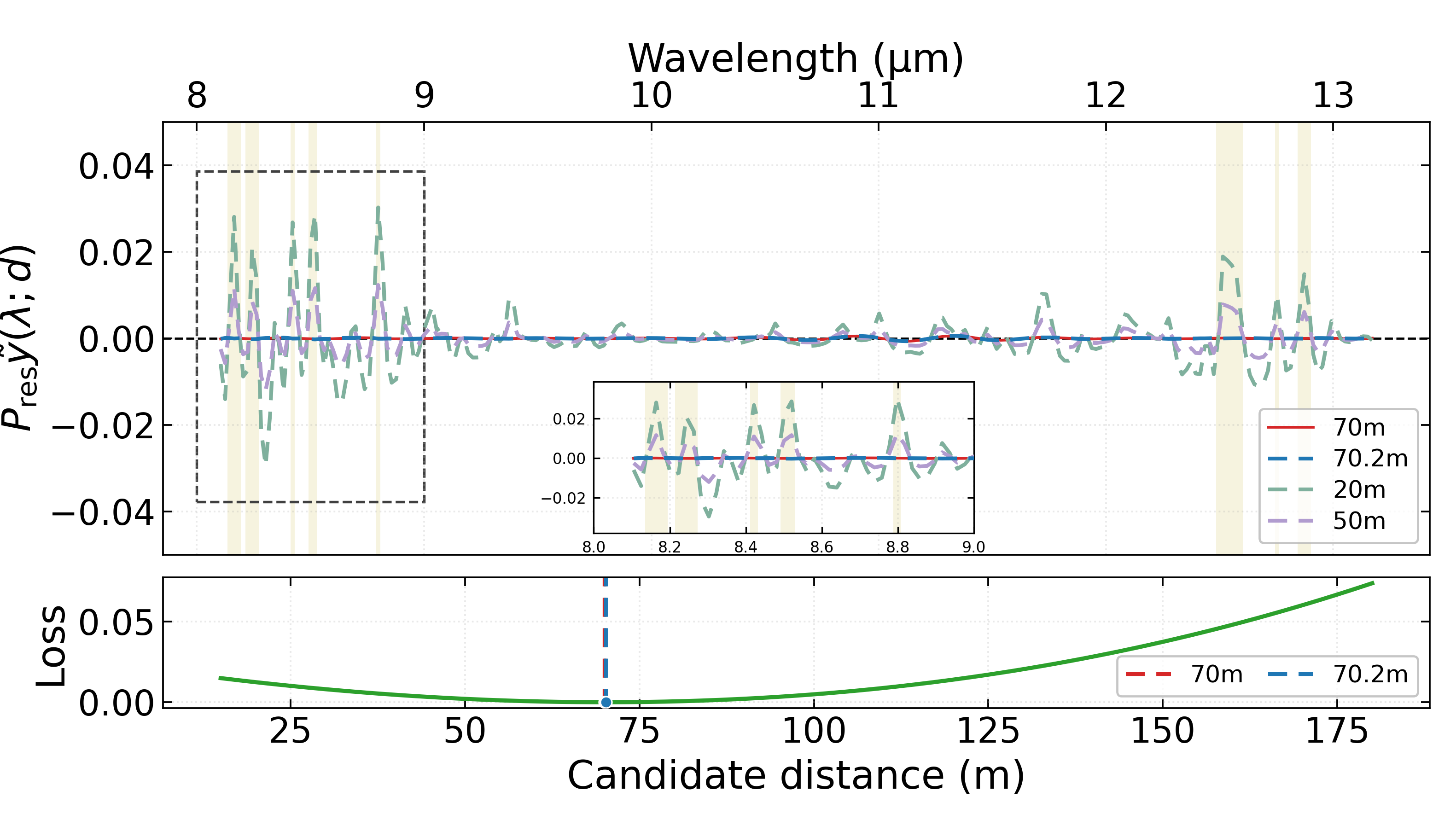}
		\label{fig:fig43b}
	}
	
	\vspace{-0.2em}
	\caption{Illustration of the distance-decoupling mechanism using a simulated rock radiance spectrum. 
		Strong absorption bands are highlighted by shaded background regions. 
		(a) Observed radiance and compensated spectra under different candidate distances. 
		When the candidate distance approaches the true distance of \(70~\mathrm{m}\), the compensated radiance becomes smoother. 
		(b) The upper panel shows the absorption residual curves under different candidate distances, while the lower panel shows the residual energy as a function of candidate distance. 
		The residual energy reaches its minimum near the true distance.}
	\label{fig:decoupled_estimation_mechanism}
	\vspace{-0.6em}
\end{figure}

In this way, the compensated radiance is decomposed into an absorption residual component and a smooth background component.
The first component is mainly related to distance, while the second component is mainly associated with temperature and emissivity. 
When the candidate distance is correct, the compensated spectrum should be smoother, and its absorption residual component should be smaller. 
Thus, distance estimation can be formulated as a one-dimensional search problem:
\begin{equation}
	\hat{d}
	=
	\arg\min_{d\in\mathcal{D}}
	\left\|
	P_{\mathrm{res}}\widetilde{\mathbf{y}}(d)
	\right\|_2^2 ,
	\label{eq:d_decoupled}
\end{equation}
where \(\mathcal{D}\) denotes the candidate distance range. 
This formulation avoids complex joint optimization over temperature, emissivity, and distance, and instead decouples distance estimation from the observed radiance.
Fig.~\ref{fig:decoupled_estimation_mechanism} illustrates this mechanism using a simulated rock spectrum with a true distance of \(70~\mathrm{m}\) and a target temperature of \(287~\mathrm{K}\), which is \(3~\mathrm{K}\) lower than the air temperature. 
In the noise-free case, the distance-decoupled estimator obtains \(70.2~\mathrm{m}\).

	The above distance-decoupled formulation relies on separating the smooth target self-emission component from the sharper distance-dependent atmospheric structures through the residual projection operator.
	However, although the material emissivity \(\varepsilon(\lambda)\) is represented by B-spline basis functions, the Planck-modulated target self-emission \(\varepsilon(\lambda)B(\lambda;T)\) does not strictly remain in the same B-spline subspace.
	Consequently, \(P_{\mathrm{res}}\widetilde{\mathbf{y}}(d)\) may retain a small out-of-subspace target component in addition to the distance-related absorption residual component.
	To quantify the distance bias attributable specifically to this projection approximation, each material emissivity spectrum is first represented in the same B-spline subspace used by ADER, after which Planck modulation is applied at different target temperatures.
	Across the 20 tested material--temperature conditions, the resulting absolute range deviation has a mean of \(0.230~\mathrm{m}\) and a maximum of \(0.886~\mathrm{m}\).
	Moreover, for a \(10~\mathrm{m}\) candidate-distance offset, the distance-mismatch-induced change in the projected residual is, on average, \(127.2\times\) the model-approximation residual.
	Together, these results support the key approximation underlying the distance-decoupled estimator: when the out-of-subspace target component is small relative to the residual variation induced by distance mismatch, the location of the residual minimum remains governed primarily by distance-dependent atmospheric absorption structures.

	\color{black}
		Having verified that the residual-projection approximation has only a limited influence on distance estimation, we further examine noise propagation in the distance-decoupled formulation.
		Transmittance compensation makes the propagated noise variance dependent on the candidate distance, causing the unweighted residual to exhibit a statistical tendency toward shorter distance estimates.
		We therefore further evaluate a covariance-weighted objective that accounts for the propagated noise variance.

		Covariance weighting effectively reduces the short-range bias induced by propagated noise under matched-model simulations.
		However, across five evaluated real-scene regions, covariance weighting reduces the absolute regional mean-distance deviation from the LiDAR reference in only one region, while increasing it in the other four regions.
		This indicates that its advantage under ideal matched-model conditions does not consistently translate into improved real-scene performance.
		The discrepancy may be associated with nonideal factors such as real sensor noise and residual model--data mismatch.
		Based on the real-scene experimental results, ADER retains the unweighted residual objective as the default distance criterion and treats the covariance-weighted formulation as a sensitivity analysis of the propagated-noise effect.
	\color{black}

	Beyond examining the bias and robustness of the distance objective, we further characterize the local precision attainable by the distance-decoupled estimator in the presence of unknown target temperature and emissivity.
	This theoretical precision analysis is derived under the simplified radiative model for emission-dominant pixels; therefore, it characterizes the local precision limit of the distance-decoupled estimator rather than that of the reflection-compensated refinement model used for reflection-dominant pixels.
	Detailed derivations and controlled validations of these properties, including the projection-approximation analysis, covariance-weighting analysis, Fisher-information derivation, Cram\'er--Rao lower-bound analysis, and Monte Carlo experiments, are provided in Appendix~B of the supplementary material.

	Once distance has been determined by the distance-decoupled estimator, the corresponding smooth compensated spectrum can then be used to estimate target temperature and recover the B-spline emissivity representation.
	Because the present work focuses on passive ranging, these subsequent temperature and emissivity estimates are not included in the main ranging evaluation; their estimation procedure and representative results are provided in Appendix~C of the supplementary material.

\subsection{Reflection-Compensated Refinement}
\label{subsec:reflective_pixel_refinement}

The distance-decoupled estimator relies on the weak-reflection assumption and is mainly applicable to emission-dominant pixels. 
For reflection-dominant pixels, neglecting downwelling radiance may cause the reflected component to be incorrectly interpreted as a longer propagation distance, leading to distance overestimation. 
Therefore, for the reflection-dominant pixel set \(\mathcal{P}_{\mathrm{ref}}\) identified by pixel classification based on downwelling-radiance cues, we further introduce the downwelling-radiance dictionary coefficient vector \(\mathbf{v}\) into the complete radiative model for distance refinement.

According to the complete radiative model in Eq.~\eqref{eq:standard_radiative_model}, reflection-dominant pixels are refined by minimizing the discrepancy between the observed spectrum and the reconstructed spectrum.
For each pixel in \(\mathcal{P}_{\mathrm{ref}}\), the refinement problem is formulated as
\begin{equation}
	\begin{aligned}
		(d^{\ast},T^{\ast},\mathbf{u}^{\ast},\mathbf{v}^{\ast})
		&=
		\arg\min_{d,T,\mathbf{u},\mathbf{v}}
		\left\|
		\hat{\mathbf{y}}(d,T,\mathbf{u},\mathbf{v})
		-
		\mathbf{y}
		\right\|_2^2,\\
		\mathrm{s.t.}\quad
		&0\leq d\leq d_{\max},\qquad
		T_{\min}\leq T\leq T_{\max},\\
		&\mathbf{0}\preceq\Phi\mathbf{u}\preceq\mathbf{1},\qquad
		v_n\geq0,\quad n=1,\ldots,N_s,\\
		&\sum_{n=1}^{N_s}v_n\leq1 .
	\end{aligned}
	\label{eq:reflective_refine_loss}
\end{equation}
Here, \(d\), \(T\), \(\mathbf{u}\), and \(\mathbf{v}=[v_1,\ldots,v_{N_s}]^{\top}\) denote the distance, temperature, B-spline emissivity control-point vector, and downwelling-radiance coefficient vector of the current pixel, respectively.
The vectors \(\mathbf{y}\) and \(\hat{\mathbf{y}}(d,T,\mathbf{u},\mathbf{v})\) denote the observed and reconstructed radiance over the operating bands, while \(\Phi\mathbf{u}\) denotes the corresponding B-spline emissivity.
The coefficient \(v_n\) is the weight of the \(n\)-th zenith-angle downwelling-radiance spectrum.
Equation~\eqref{eq:reflective_refine_loss} is solved using Adam independently for each pixel in \(\mathcal{P}_{\mathrm{ref}}\)~\cite{kingma_adam_2015}.
\textcolor{black}{
	After each Adam update, the distance,
	temperature, and B-spline control coefficients
	are clipped to their prescribed bounds.
	The downwelling-radiance coefficients are
	clamped to be nonnegative and rescaled
	when their sum exceeds unity, thereby
	enforcing the constraints in
	Eq.~\eqref{eq:reflective_refine_loss}.
}

Through this pixel-classification strategy, ADER applies one-dimensional distance-decoupled estimation to most emission-dominant pixels and introduces reflection-compensated refinement only for a small number of reflection-dominant pixels.
This avoids discarding the distance estimates of reflection-dominant regions while reducing the computational cost of full-image joint optimization. 
The correction effect on real reflection-dominant regions is further evaluated in the experimental section.


\section{Greedy Band Selection Based on Multi-Scene Effective Fisher Information for Distance}
\label{sec:band_selection}

The analysis in Section~IV indicates that distance information is unevenly distributed across wavelengths and is concentrated mainly in atmospheric absorption regions, whereas full-band hyperspectral observations contain substantial spectral redundancy. 
To enable efficient few-band ranging, we use Fisher information to quantify the distance sensitivity of individual bands and construct a compact wavelength subset based on multi-scene effective Fisher information for distance. 
Unlike feature extraction methods, band selection preserves the original spectral channels and thus retains direct wavelength-level physical interpretability~\cite{bioucas-dias_hyperspectral_2013,sun_hyperspectral_2019}.
To remain consistent with the experimental scene, the atmospheric transmittance prior associated with the main ranging scene is used throughout the following analysis~\cite{gallastegi_ozone_2026}.

\subsection{Fisher Information Analysis}
\label{subsec:fisher_information_analysis}

Because band selection primarily serves the distance-decoupled estimator for emission-dominant pixels, band sensitivity is evaluated using the radiative model in Eq.~\eqref{eq:high_emissivity_model}. 
The noise-free mean radiance at the \(k\)-th band is denoted by
\(\mu_k(\boldsymbol{\theta})\), where
\(\boldsymbol{\theta}=[d,T,\mathbf{u}^{\top}]^{\top}\).
For the \(c\)-th reference scene, the derivatives are evaluated at
\[
\boldsymbol{\theta}^{(c)}
=
[d^{(c)},T^{(c)},(\mathbf{u}^{(c)})^{\top}]^{\top}.
\]
The atmospheric transmittance prior is fixed to that associated with the main ranging scene. Other prescribed scene-level quantities are treated as known reference conditions and are omitted from \(\boldsymbol{\theta}^{(c)}\) for brevity.
Assuming independent zero-mean Gaussian noise with a common variance
\(\sigma^2\) across bands, the Fisher information contributed by a band is proportional to the squared derivative of its mean radiance with respect to the parameter of interest~\cite{kay_fundamentals_1993}.
Accordingly, the single-band distance and temperature scores are defined as
\begin{equation}
	\begin{aligned}
		S_d(k;c)
		&=
		\frac{1}{\sigma^2}
		\left.
		\left(
		\frac{\partial\mu_k(\boldsymbol{\theta})}{\partial d}
		\right)^2
		\right|_{\boldsymbol{\theta}=\boldsymbol{\theta}^{(c)}},
		\\
		S_T(k;c)
		&=
		\frac{1}{\sigma^2}
		\left.
		\left(
		\frac{\partial\mu_k(\boldsymbol{\theta})}{\partial T}
		\right)^2
		\right|_{\boldsymbol{\theta}=\boldsymbol{\theta}^{(c)}} .
	\end{aligned}
	\label{eq:single_band_fisher_score}
\end{equation}
These scores have a direct interpretation: a larger value indicates greater local sensitivity of the radiance to the corresponding parameter.

Because target distance, temperature, emissivity, and other radiative conditions vary across scenes, a score evaluated under a single reference scene may be scene specific. 
We therefore average the single-band scores over a set
\(\mathcal{C}\) of representative reference scenes:
\begin{equation}
	\overline{S}_d(k)
	=
	\frac{1}{|\mathcal{C}|}
	\sum_{c\in\mathcal{C}}
	S_d(k;c),
	\qquad
	\overline{S}_T(k)
	=
	\frac{1}{|\mathcal{C}|}
	\sum_{c\in\mathcal{C}}
	S_T(k;c).
	\label{eq:mean_single_band_fisher_score}
\end{equation}
Here, \(\mathcal{C}\) denotes the reference-scene set, while
\(\overline{S}_d(k)\) and \(\overline{S}_T(k)\) denote the multi-scene average distance and temperature scores of the \(k\)-th band, respectively.
	Specifically, \(\mathcal{C}\) contains 75 reference conditions formed by
	\(T_{\mathrm{air}}\in\{285,290,295\}~\mathrm{K}\),
	\(\Delta T=T-T_{\mathrm{air}}\in\{-8,-4,4,8,15\}~\mathrm{K}\), and
	\(d\in\{20,40,60,80,120\}~\mathrm{m}\).
	All reference conditions use the same scene-level atmospheric transmittance prior.
	Averaging over these conditions reduces the dependence of band selection on a single nominal distance or temperature contrast.

\begin{figure}[!t]
	\centering
	\subfloat[Single-band Fisher score]{
		\includegraphics[width=0.95\columnwidth]{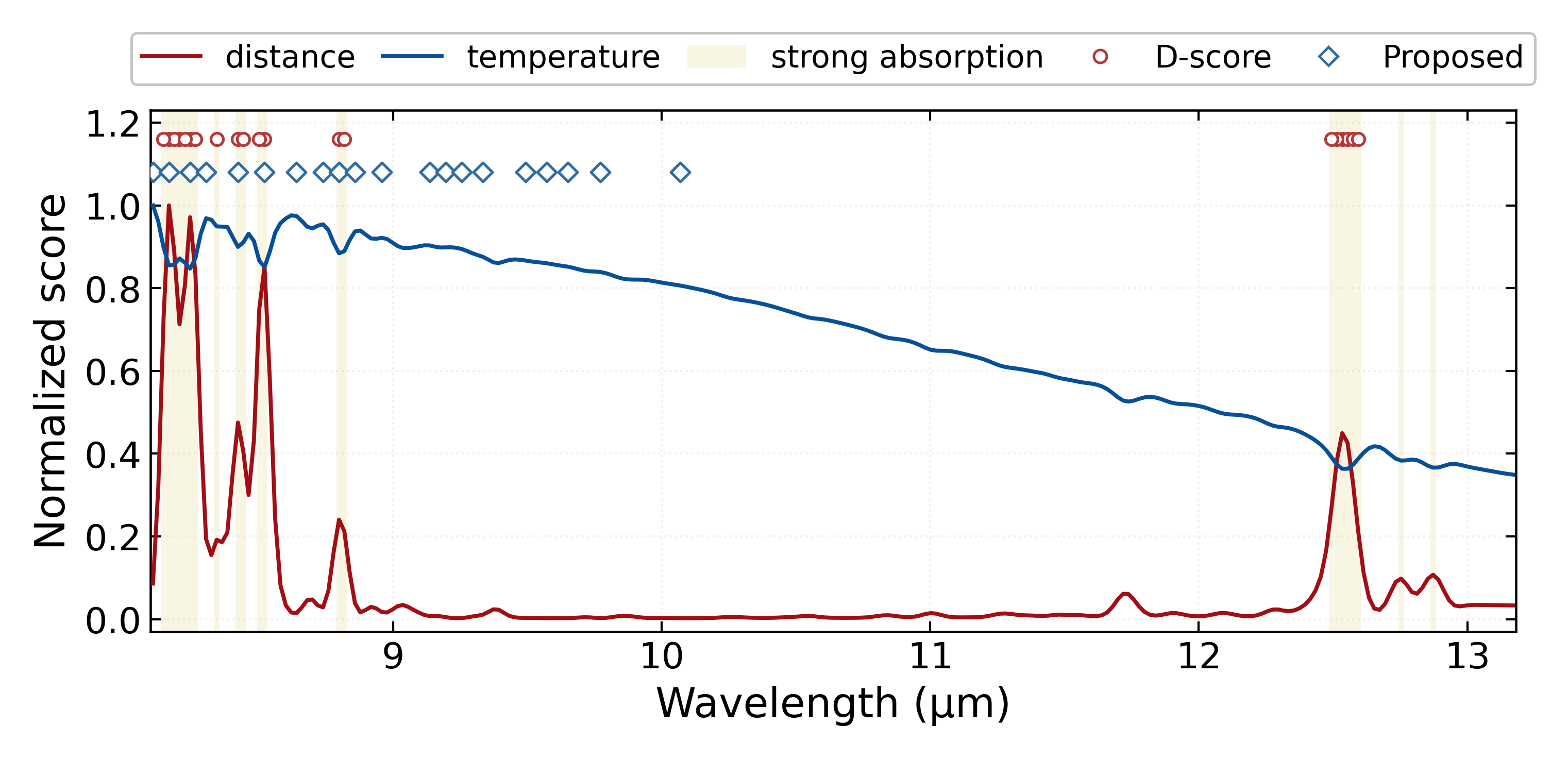}
		\label{fig:fig51a}
	}\\[2pt]
	\subfloat[Effective Fisher information for distance]{
		\includegraphics[width=0.95\columnwidth]{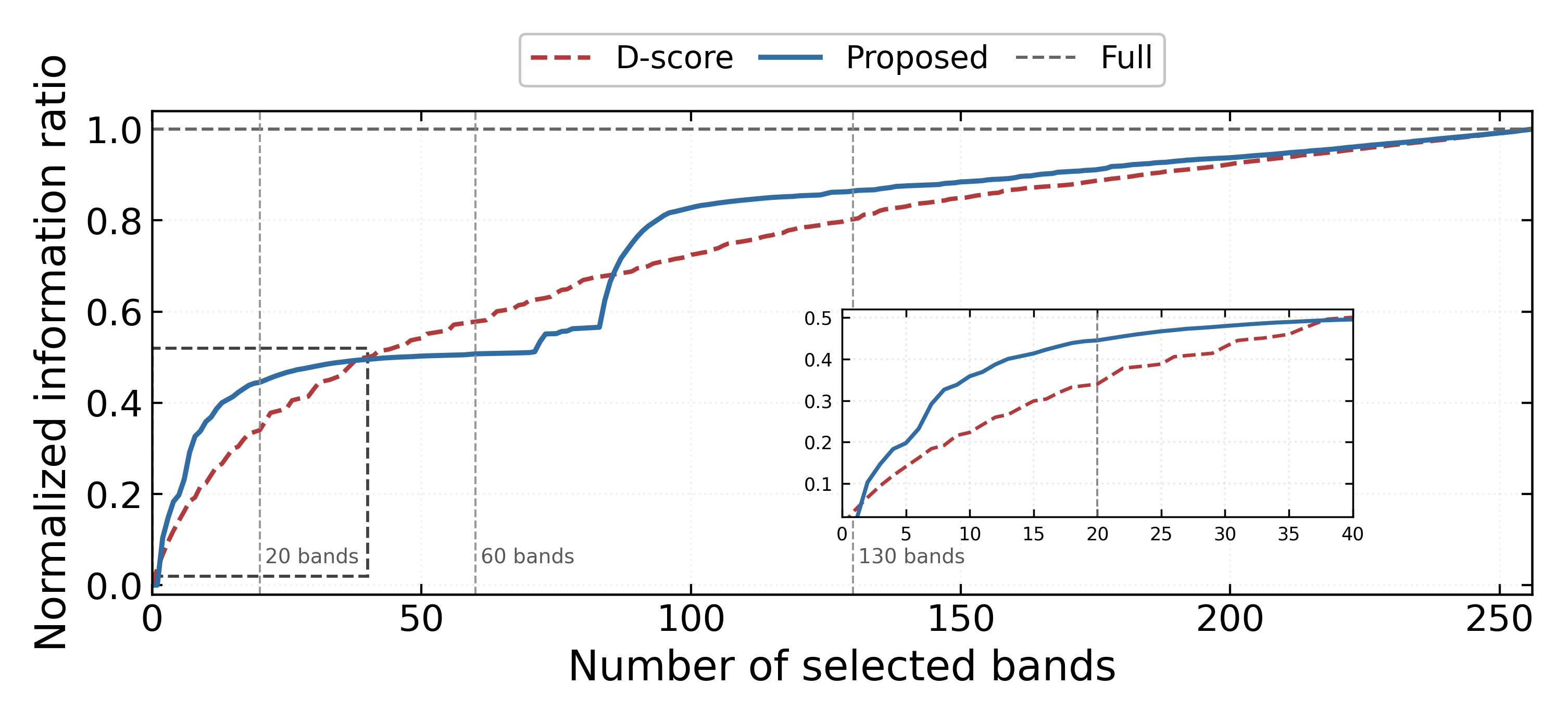}
		\label{fig:fig51b}
	}
	\caption{Fisher-information-based band selection analysis.
		(a) Single-band Fisher scores averaged over simulated rock spectra under different reference scenes.
		Each curve is independently normalized to compare the information distribution across wavelengths.
		The markers denote the bands selected by direct distance-score ranking and by the proposed greedy band selection strategy, respectively.
		(b) Effective Fisher information for distance under different band selection strategies, normalized by the full-band value.}
	\label{fig:fig51}
\end{figure}

Fig.~\ref{fig:fig51}(a) shows the averaged single-band distance and temperature scores. 
The distance score is strongly nonuniform and peaks in the main atmospheric absorption regions, whereas the temperature score varies more smoothly with wavelength. 
This contrast indicates that a limited number of absorption-sensitive bands contain a substantial portion of the distance-related information and provides the basis for few-band ranging.
	Accordingly, bands within or near pronounced atmospheric absorption structures are favored because their transmittance exhibits stronger distance-dependent modulation.
	The final subset is not simply composed of the deepest absorption bands, because the effective Fisher criterion also accounts for temperature--emissivity coupling and inter-band redundancy, thereby favoring complementary spectral information for distance estimation.

\subsection{Greedy Incremental Band Selection}
\label{subsec:greedy_incremental_band_selection}

A straightforward strategy is to rank the bands independently according to
\(\overline{S}_d(k)\).
However, independent ranking ignores inter-band correlation and may repeatedly select bands containing similar distance information~\cite{sun_hyperspectral_2019,zhao_spectral_spatial_ga_2021,zhang_joint_graph_selfrep_2024}. 
We therefore construct the band subset greedily according to the effective distance information accumulated by the currently selected bands.

For the \(k\)-th band under the \(c\)-th reference scene, the local response to the physical parameters is represented by
\begin{equation}
	\mathbf{g}_k^{(c)}
	=
	\left.
	\left[
	\frac{\partial\mu_k(\boldsymbol{\theta})}{\partial d},
	\frac{\partial\mu_k(\boldsymbol{\theta})}{\partial T},
	\frac{\partial\mu_k(\boldsymbol{\theta})}{\partial u_1},
	\cdots,
	\frac{\partial\mu_k(\boldsymbol{\theta})}{\partial u_M}
	\right]^{\top}
	\right|_{\boldsymbol{\theta}=\boldsymbol{\theta}^{(c)}} .
	\label{eq:band_gradient_vector}
\end{equation}

For a selected band subset \(\Omega\), the multi-scene accumulated Fisher information matrix is defined as
\begin{equation}
	\overline{\mathbf{F}}(\Omega)
	=
	\frac{1}{|\mathcal{C}|}
	\sum_{c\in\mathcal{C}}
	\frac{1}{\sigma^2}
	\sum_{k\in\Omega}
	\mathbf{g}_k^{(c)}
	\left(\mathbf{g}_k^{(c)}\right)^{\top}.
	\label{eq:multi_scene_accumulated_fim}
\end{equation}
The Fisher information of the selected subset is first accumulated within each reference scene and then averaged over the representative scene set. Unlike the independent single-band scores, 
\(\overline{\mathbf{F}}(\Omega)\)  depends on the current band subset and describes the average accumulated constraint provided by this subset on the unknown parameters across the reference scenes.

The nuisance parameters are collected in
\(\boldsymbol{\xi}=[T,\mathbf{u}^{\top}]^{\top}\).
Accordingly, the accumulated Fisher information matrix is partitioned as
\begin{equation}
	\overline{\mathbf{F}}(\Omega)
	=
	\begin{bmatrix}
		\overline{F}_{dd}(\Omega)
		&
		\overline{\mathbf{F}}_{d\xi}(\Omega)
		\\
		\overline{\mathbf{F}}_{\xi d}(\Omega)
		&
		\overline{\mathbf{F}}_{\xi\xi}(\Omega)
	\end{bmatrix}.
	\label{eq:multi_scene_fim_partition}
\end{equation}
After accounting for the coupling of distance with temperature and emissivity, the multi-scene effective Fisher information for distance is defined as
\begin{equation}
	\overline{J}_d(\Omega)
	=
	\overline{F}_{dd}(\Omega)
	-
	\overline{\mathbf{F}}_{d\xi}(\Omega)
	\left(
	\overline{\mathbf{F}}_{\xi\xi}(\Omega)
	\right)^{\dagger}
	\overline{\mathbf{F}}_{\xi d}(\Omega).
	\label{eq:multi_scene_effective_distance_fisher}
\end{equation}
The generalized Schur-complement term removes the portion of distance information coupled to uncertainty in temperature and emissivity. 
The Moore--Penrose pseudoinverse is used because
\(\overline{\mathbf{F}}_{\xi\xi}(\Omega)\) may be rank deficient or ill-conditioned when only a few bands are selected.
	In the numerical evaluation, the pseudoinverse is computed using
	\(r_{\mathrm{cond}}=10^{-15}\), which is consistent with the numerical truncation scale commonly used in double-precision SVD computations.
	Since the Schur complement is a nonlinear operation, averaging the Fisher information matrices over different reference conditions before applying the Schur complement is generally not equivalent to applying the Schur complement separately under each reference condition and then averaging the resulting effective information.
	
	Different pixels in a real scene generally correspond to different target distances and temperatures.
	We therefore use a set of representative reference conditions to simulate such pixel-state variations and, under a fixed scene-level atmospheric transmittance prior, design one common band subset for these conditions to reduce the dependence of band selection on any single reference condition.
	Specifically, the Fisher information matrices are first averaged over the reference conditions to integrate the parameter sensitivities and coupling relationships of distance, temperature, and emissivity under different conditions; the Schur complement is then applied to the averaged information matrix to remove the component of distance information coupled with temperature and emissivity, thereby yielding the effective distance information used for band selection.
	Analysis and comparison of the alternative information-aggregation order are provided in Supplementary Appendix~F-A.

The multi-scene effective Fisher information
\(\overline{J}_d(\Omega)\) is used as the main criterion for band selection.
Since it depends on the currently selected band subset, it accounts for the redundancy among different bands.
Exhaustive enumeration of all possible subsets is computationally prohibitive, and similar measurement-subset selection problems are commonly addressed using approximate or greedy strategies~\cite{joshi_sensor_2009,krause_near-optimal_2008}.

Given the currently selected subset \(\Omega_t\), the effective distance-information gain of a candidate band \(k\notin\Omega_t\) is defined as
\begin{equation}
	\Delta\overline{J}_d(k\mid\Omega_t)
	=
	\overline{J}_d(\Omega_t\cup\{k\})
	-
	\overline{J}_d(\Omega_t).
	\label{eq:multi_scene_incremental_gain}
\end{equation}
At each step, the band with the largest
\(\Delta\overline{J}_d(k\mid\Omega_t)\) is selected and added to the subset.

In addition, excessively large wavelength gaps between the selected bands may destabilize the B-spline emissivity fitting and affect the ranging results.
Therefore, a wavelength-gap penalty is introduced, and the band selected at each step is determined by
\begin{equation}
	k_{t+1}
	=
	\arg\max_{k\notin\Omega_t}
	\left[
	\Delta\overline{J}_d(k\mid\Omega_t)
	-
	\lambda_{\mathrm{gap}}
	P_{\mathrm{gap}}(k,\Omega_t)
	\right],
	\label{eq:gap_regularized_incremental_selection}
\end{equation}
followed by
\begin{equation}
	\Omega_{t+1}
	=
	\Omega_t\cup\{k_{t+1}\},
	\label{eq:selected_band_set_update}
\end{equation}
	\textcolor{black}{
		Here, the wavelength-gap penalty is defined as
	}
\begin{equation}
	\color{black}
	P_{\mathrm{gap}}(k,\Omega_t)
	=
	\left[
	\max\left(0,
	\min_{i\in\Omega_t}|\lambda_k-\lambda_i|
	-2\Delta\lambda_{\mathrm{avg}}
	\right)
	\right]^2,
	\label{eq:wavelength_gap_penalty}
\end{equation}
\textcolor{black}{
	where \(\Delta\lambda_{\mathrm{avg}}
	=(\lambda_{\max}-\lambda_{\min})/(N_b-1)\)
	is the nominal average wavelength spacing.
	The penalty is set to zero for an empty
	subset, and \(\lambda_{\mathrm{gap}}=0.20\).
}
	The greedy search starts from an empty subset, \(\Omega_0=\emptyset\).
	Candidate bands are evaluated in ascending wavelength order, and exact ties retain the smaller-index candidate.

	The final \(N_b=20\) band subset used in the ranging experiments is
	\begin{equation*}
		\begin{aligned}
			\{&
			8.0443,\ 8.1037,\ 8.1630,\ 8.2422,\ 8.3016,\ 8.4204,\ 8.5194,\\
			&
			8.6185,\ 8.7374,\ 8.7969,\ 8.8564,\ 8.9555,\ 9.1341,\ 9.1936,\\
			&
			9.2532,\ 9.3325,\ 9.5112,\ 9.5708,\ 9.6502,\ 9.7694
			\}~\mu\mathrm{m}.
		\end{aligned}
	\end{equation*}
	Further analyses of the Fisher-information aggregation order and the numerical stability of the greedy band-selection procedure are provided in Appendix~F of the supplementary material.

Fig.~\ref{fig:fig51}(b) compares the normalized effective Fisher information for distance obtained by direct distance-score ranking and the proposed greedy strategy.
With the full-band information normalized to one, the effective distance information retained by direct ranking increases gradually with the number of selected bands.
In contrast, the proposed strategy retains a substantially larger fraction of the full-band distance information when only a small number of bands are selected and reaches a relatively high information level with about 20 bands.
Although the effective Fisher information does not directly correspond to the final ranging accuracy, it provides guidance for setting the number of selected bands in the experiments.


\section{Experiments}
\label{sec:experiments}
\begin{figure*}[!t]
	\centering
	\setlength{\abovecaptionskip}{3pt}
	\setlength{\belowcaptionskip}{0pt}
	
	\includegraphics[width=0.92\textwidth]{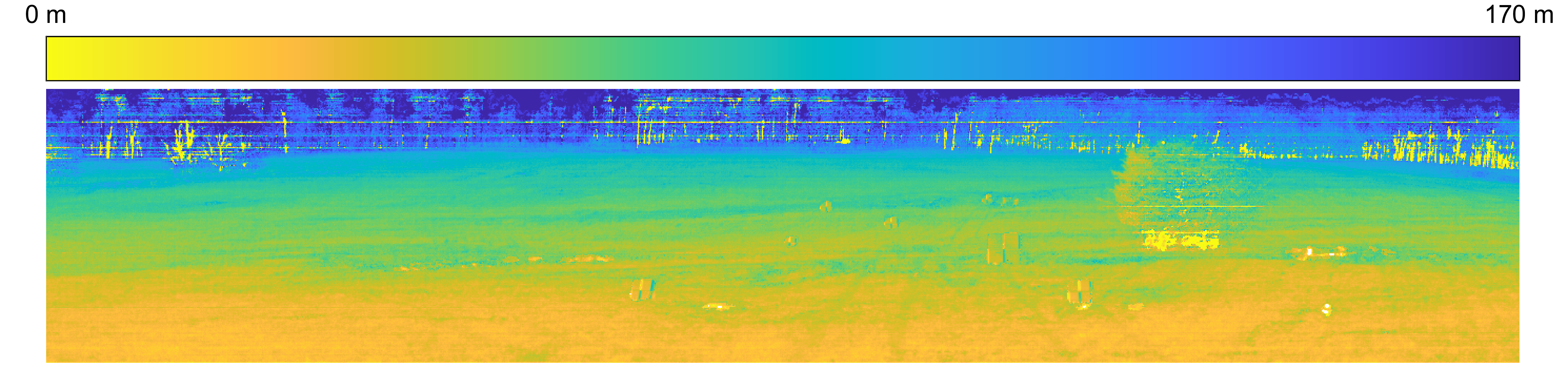}\par
	\vspace{-1mm}
	\makebox[\textwidth][c]{\footnotesize (a) ADER full-band}
	
	\vspace{1.2mm}
	
	\includegraphics[width=0.92\textwidth]{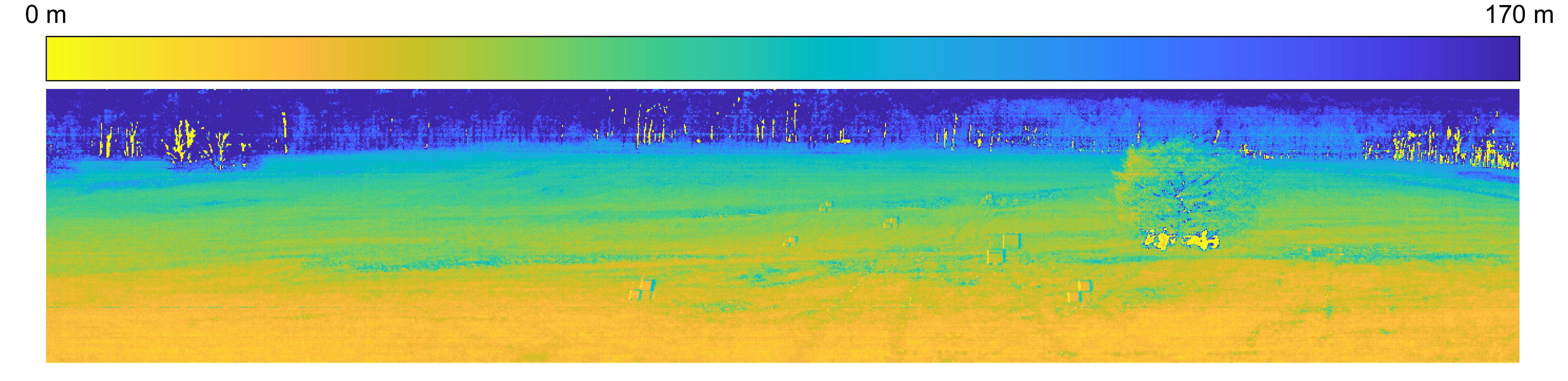}\par
	\vspace{-1mm}
	\makebox[\textwidth][c]{\footnotesize (b) ADER 20-band}
	
	\vspace{1.2mm}
	
	\includegraphics[width=0.92\textwidth]{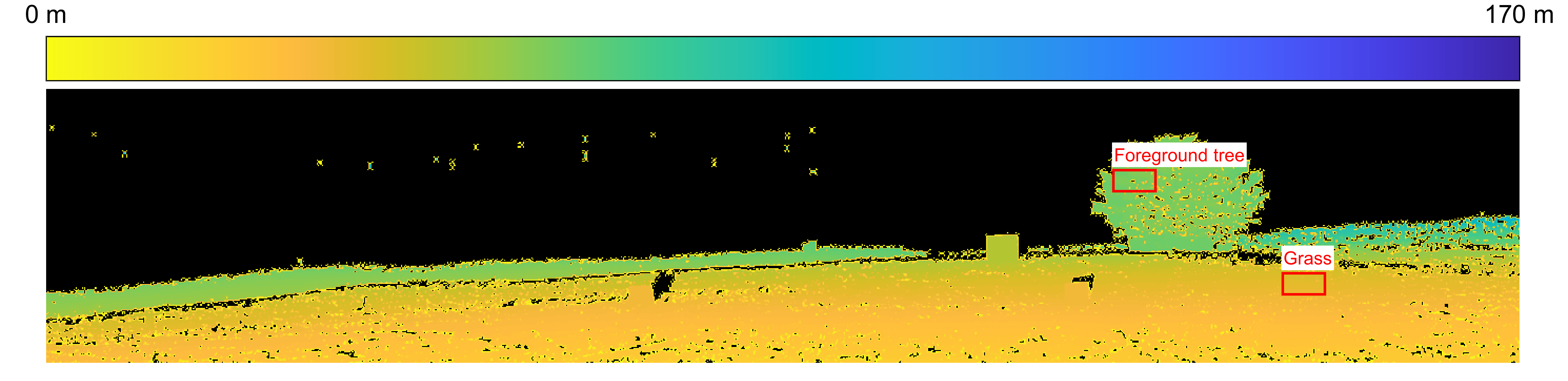}\par
	\vspace{-1mm}
	\makebox[\textwidth][c]{\footnotesize (c) LiDAR reference}
	
	\vspace{1mm}
	\caption{Real-scene ranging comparison.
		(a) ADER full-band.
		(b) ADER 20-band.
		(c) Approximately registered LiDAR map with the evaluated grass and foreground-tree regions marked.}
	\label{fig:fig61}
	
	\vspace{2.0mm}
	
	\begingroup
	\makeatletter
	\def\@captype{table}
	\makeatother
	
	\caption{Quantitative comparison on three \(20\times40\) representative patches.
		The values are reported as mean distance \(\pm\) standard deviation.
		The HS baseline (reported) is used as a literature reference, while the
		HS baseline (reproduced) and ADER results are evaluated on the same
		selected patches.}
	\label{tab:tab2}
	
	\vspace{1.2mm}
	\footnotesize
	\renewcommand{\arraystretch}{1.05}
	\setlength{\tabcolsep}{5.2pt}
	\setlength{\heavyrulewidth}{1.0pt}
	\setlength{\lightrulewidth}{0.5pt}
	
	\begin{tabular*}{0.96\textwidth}
		{@{\extracolsep{\fill}}lcccc}
		\toprule
		Method
		& Grass
		& Foreground tree
		& Background forest
		& Runtime \\
		\midrule
		
		ADER full-band
		& \(\mathbf{35.45~\mathrm{m} \pm 3.37~\mathrm{m}}\)
		& \(\mathbf{56.11~\mathrm{m} \pm 10.08~\mathrm{m}}\)
		& \(127.66~\mathrm{m} \pm \mathbf{9.54~\mathrm{m}}\)
		& \(77.2~\mathrm{s}\) \\
		
		ADER 20-band
		& \(34.85~\mathrm{m} \pm 3.98~\mathrm{m}\)
		& \(61.05~\mathrm{m} \pm 14.46~\mathrm{m}\)
		& \(135.77~\mathrm{m} \pm 9.97~\mathrm{m}\)
		& \(\mathbf{56.23~\mathrm{s}}\) \\
		
		HS baseline (reported)
		& \(48.50~\mathrm{m} \pm 4.20~\mathrm{m}\)
		& \(76.00~\mathrm{m} \pm 21.20~\mathrm{m}\)
		& \(162.30~\mathrm{m} \pm 17.80~\mathrm{m}\)
		& -- \\
		
		HS baseline (reproduced)
		& \(39.73~\mathrm{m} \pm 4.09~\mathrm{m}\)
		& \(83.67~\mathrm{m} \pm 21.51~\mathrm{m}\)
		& \(194.52~\mathrm{m} \pm 12.64~\mathrm{m}\)
		& \(167.79~\mathrm{min}\) \\
		
		LiDAR median
		& \(35.25~\mathrm{m}\)
		& \(57.48~\mathrm{m}\)
		& --
		& -- \\
		
		\bottomrule
	\end{tabular*}
	
	\vspace{-0.6em}
	\endgroup
	\end{figure*}
\subsection{Experimental Data and Runtime Settings}
\label{subsec:experimental_settings}

We evaluate ADER on LWIR hyperspectral data from the DARPA Invisible Headlights dataset~\cite{darpa_invisible_headlights_dataset,yellin_concurrent_band_2024}. 
This dataset was collected for autonomous navigation in complex off-road environments and contains synchronized observations from multiple sensing modalities, including visible imaging, hyperspectral imaging, and LiDAR. 
The recorded scenes include natural environments such as forests, sandy terrain, and grassland.
The LWIR hyperspectral data were acquired by a pushbroom LWIR hyperspectral imager equipped with a cooled HgCdTe detector. 
The two-dimensional sensor array spans the spectral dimension and the vertical spatial dimension, and the full data cube is obtained by scanning along the horizontal direction. 
This pushbroom acquisition mechanism may introduce horizontal stripe artifacts because of nonuniform pixel responses or abnormal spectral pixels.
According to the sensor parameters reported in prior work, the spectrometer has a vertical field of view (VFOV) of \(11.6^\circ\) and a horizontal field of view (HFOV) of \(57^\circ\). 
The focal length is \(50~\mathrm{mm}\), and the f-number is \(f/0.9\)~\cite{dorken_gallastegi_absorption-based_2025}. 
For each pixel in the scene, the sensor acquires 256 spectral channels from \(8.0~\mu\mathrm{m}\) to \(13.2~\mu\mathrm{m}\), with a spectral resolution of approximately \(40~\mathrm{nm}\). 
The typical sensor noise is approximately 
\(0.01~\mathrm{W}\cdot\mathrm{m}^{-2}\cdot\mathrm{sr}^{-1}\cdot\mu\mathrm{m}^{-1}\), 
corresponding to an SNR of about \(1000{:}1\) at \(10~\mu\mathrm{m}\)~\cite{dorken_gallastegi_absorption-based_2025,bao_heat-assisted_2023}. 
{\color{black}
	To reduce wavelength-calibration mismatch, we develop a scene-adaptive wavelength-correction method that performs global wavelength correction before ranging by aligning the measured and simulated atmospheric absorption structures~\cite{wang_spectral_calibration_2010,yamamoto_spectral_smile_2022}; its rationale, implementation, quantitative validation under known wavelength shifts, and ranging sensitivity are detailed in Appendix~D of the supplementary material.
}

The dataset provides LiDAR distance measurements approximately registered with the image data, which are used as reference information for evaluating passive ranging results.
We use Path5\_Step1 as the main experimental scene for passive absorption-based ranging, following prior studies on this scene~\cite{dorken_gallastegi_absorption-based_2025,gallastegi_ozone_2026}. 
This scene was collected at 19:41 local time on April 13, 2021, under low-illumination and weak-temperature-contrast conditions. 
The scene mainly consists of rolling grassy terrain, and also contains foreground trees, distant forest regions, and several checkerboard calibration targets with strong reflective properties. 
This scene contains both emission-dominant and reflection-dominant regions, making it suitable for evaluating ADER under weak temperature contrast and for analyzing pixel classification, reflection-compensated refinement, and band selection.

For the atmospheric priors, we use the atmospheric coefficients and LiDAR reference data released by prior work for this scene~\cite{gallastegi_ozone_2026}. 
We further assume that the atmospheric state is approximately uniform within the scene, and use the same unit-path transmittance and air temperature for all pixels. 
The LiDAR distance map and the LWIR hyperspectral image are only approximately registered and do not have strict pixel-wise correspondence.
Therefore, ranging performance is evaluated using local region statistics, distance profiles, and cross-scene visual comparisons rather than full-image pixel-wise error metrics.
	Residual registration uncertainty may still cause the selected hyperspectral and LiDAR regions to cover slightly different physical areas, particularly near object boundaries and abrupt depth discontinuities. 
	To minimize the influence of such spatial mismatch on quantitative ranging evaluation, the evaluation regions are selected away from obvious object boundaries and large depth gradients whenever possible. 
	Specifically, the grass and checkerboard evaluation regions are located within the interiors of the corresponding regions, while the foreground-tree evaluation region is selected without crossing the tree--background boundary.
For patch-level quantitative evaluation, the median distance of the corresponding LiDAR patch is used as the reference distance to reduce the influence of local registration mismatch and sparse invalid LiDAR samples.
	Appendix~E of the supplementary material quantifies the sensitivity of these regional statistics to registration offsets.

The proposed method is implemented on a platform equipped with an AMD Ryzen 7 5800H CPU, an NVIDIA GeForce RTX 3060 GPU with 6 GB memory, and 16 GB RAM. 
All runtime measurements are recorded under the same input data and atmospheric priors, and include only the main distance estimation process, excluding image visualization and error-statistics computation. 
For comparison, the public hyperspectral ranging code from prior work is run on the same scene and the same experimental platform, and its runtime is recorded as the computational-efficiency reference~\cite{gallastegi_ozone_2026}. 
Unless otherwise specified, all experiments use fixed implementation parameters.
The main implementation settings and reproducibility details are provided in Appendix A of the supplementary material.
The sensitivity of the B-spline control-point number and the robustness of ozone-feature-based pixel classification are analyzed in Appendix E of the supplementary material.
\textcolor{black}{
	Appendix E further evaluates how controlled pixel-classification-mask perturbations propagate to the final ranging results.
}
\textcolor{black}{
	Detailed stage-wise runtime breakdowns and cross-GPU comparisons are provided in Appendix G-A of the supplementary material.
}

\subsection{Real-Scene Ranging Results}
\label{subsec:real_scene_ranging_results}

\begin{figure}[!t]
	\centering
	\subfloat[Normalized effective distance Fisher information.]{
		\includegraphics[width=\columnwidth]{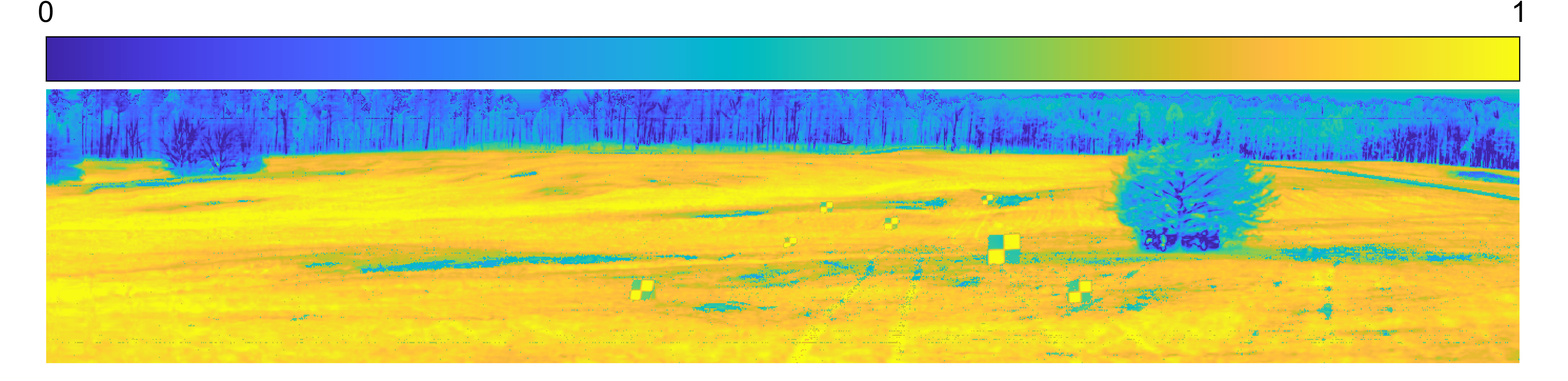}
		\label{fig:fig61_observability_a}
	}\\[2pt]
	\subfloat[Distance map after low-information pixel flagging.]{
		\includegraphics[width=\columnwidth]{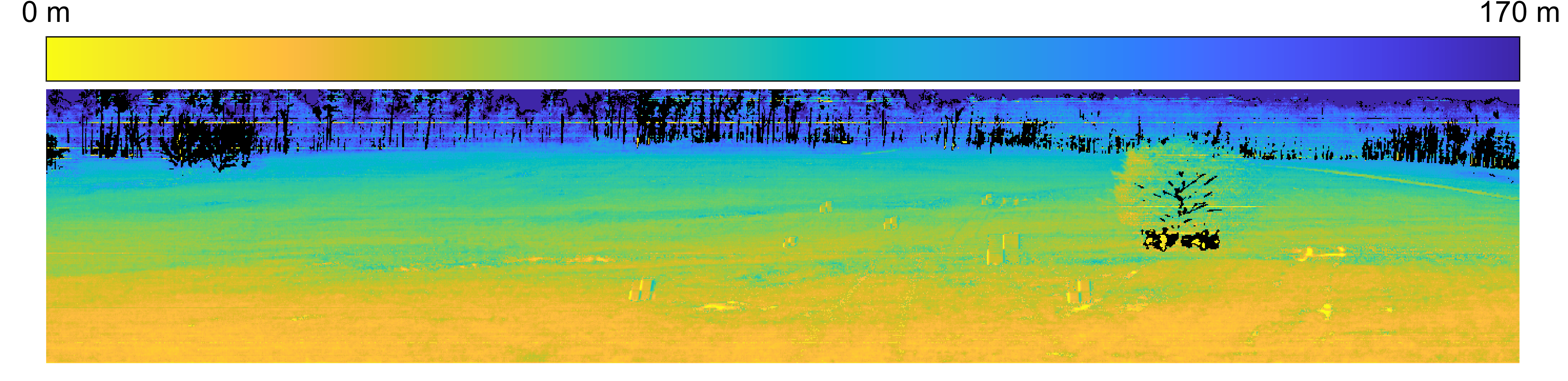}
		\label{fig:fig61_observability_b}
	}
	\caption{Real-scene distance-observability and reliability analysis.
			(a) Normalized local effective distance Fisher information, where larger values indicate higher distance observability.
			(b) Full-scene distance map after flagging low-information pixels using the scene-adaptive threshold derived from the Cramér--Rao lower bound.}
	\label{fig:fig61_observability}
\end{figure}

\begin{figure*}[!t]
	\centering
	\setlength{\abovecaptionskip}{2pt}
	\setlength{\belowcaptionskip}{0pt}
	
	\subfloat[Decoupled only]{
		\includegraphics[width=0.31\textwidth]{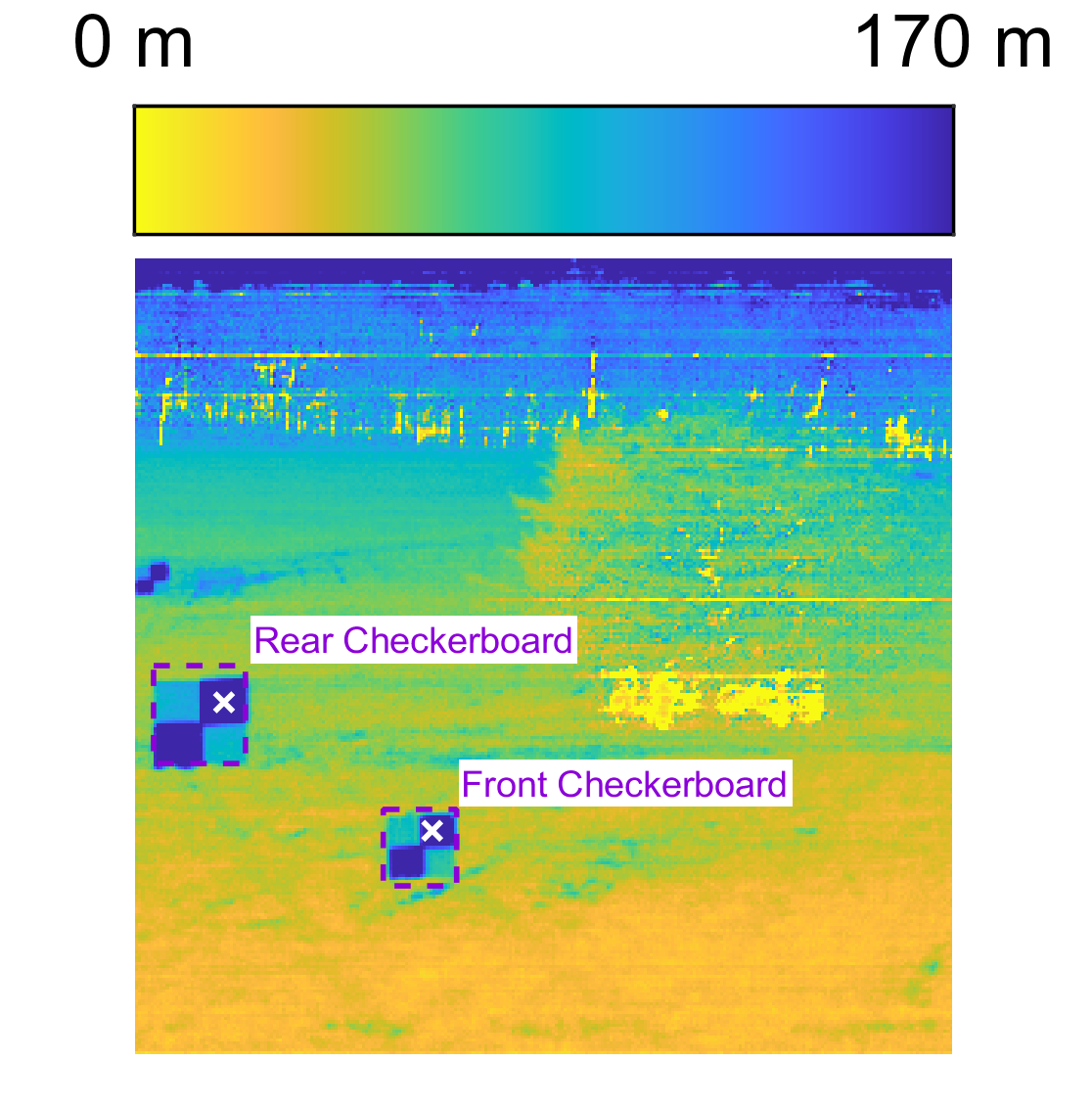}
		\label{fig:fig62a}
	}
	\hfill
	\subfloat[ADER full-band]{
		\includegraphics[width=0.31\textwidth]{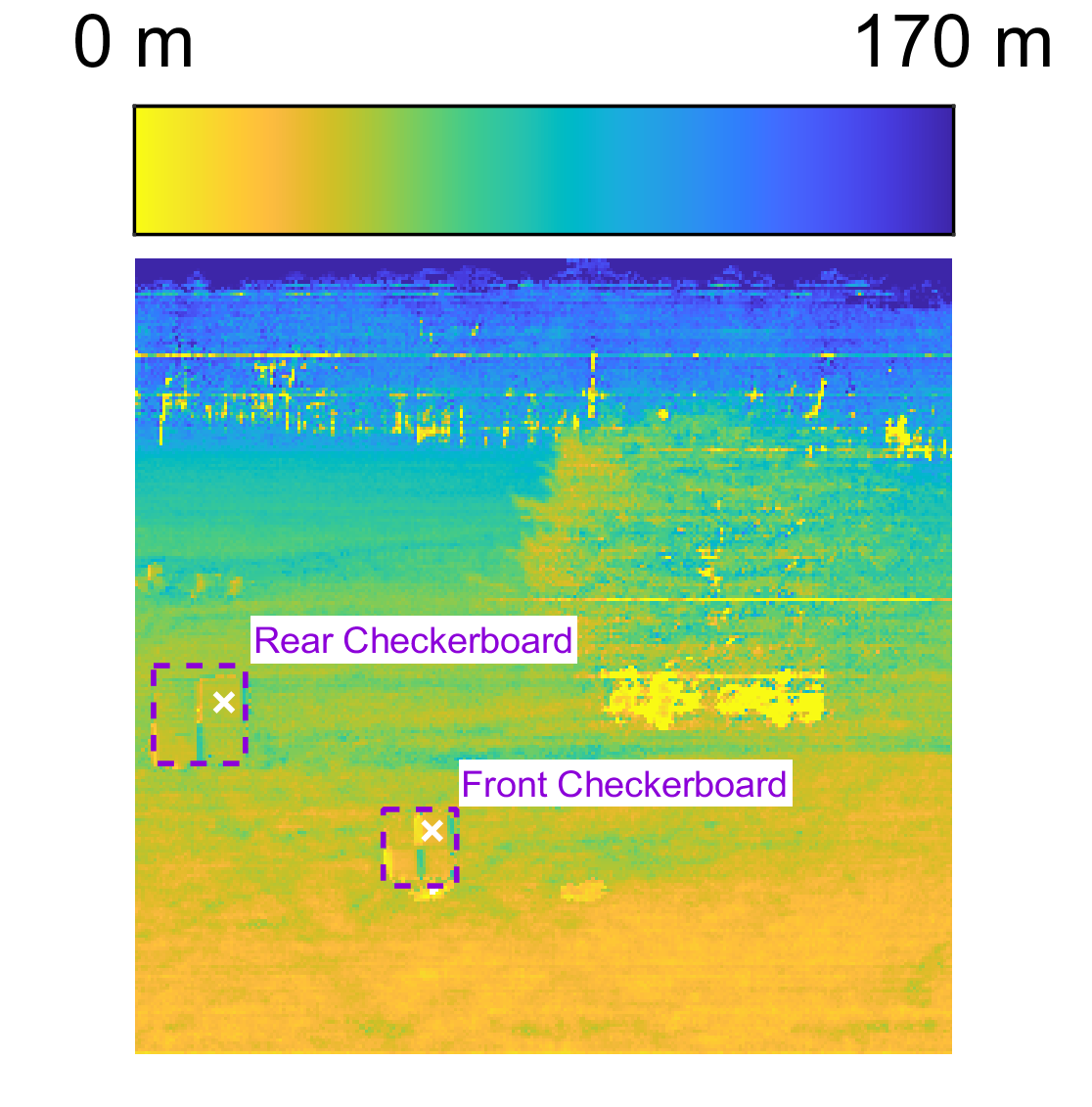}
		\label{fig:fig62b}
	}
	\hfill
	\subfloat[RGB reference]{
		\includegraphics[width=0.31\textwidth]{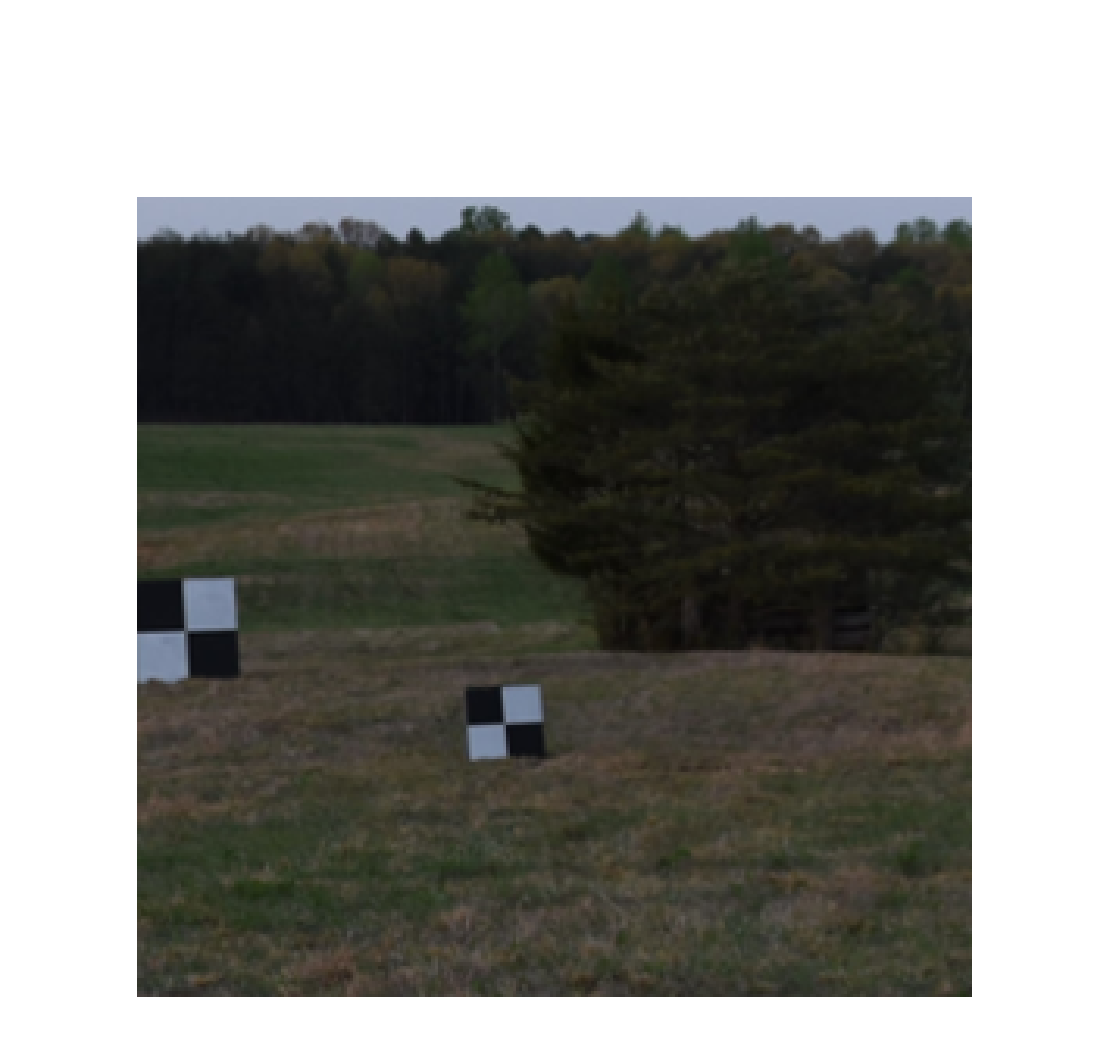}
		\label{fig:fig62c}
	}

	\vspace{-0.2em}
	\caption{Reflection-compensated refinement on reflective boards.
		(a) Decoupled-only result, where the reflective boards inside the dashed boxes are overestimated.
		(b) ADER full-band after refinement.
		(c) RGB reference.
		White crosses mark the centers of the two \(8\times8\) evaluation patches.}
	\label{fig:fig62}
	\vspace{-0.6em}
\end{figure*}

Fig.~\ref{fig:fig61} compares the ranging results on a real LWIR hyperspectral scene. 
Fig.~\ref{fig:fig61}(a) presents the ADER full-band result using 256 bands, Fig.~\ref{fig:fig61}(b) presents the result using the selected 20 bands, and Fig.~\ref{fig:fig61}(c) presents the approximately registered LiDAR reference. 
The full-band result shows a spatial distribution consistent with the LiDAR reference. 
When only the selected 20 bands are used, the ranging result remains close to the full-band result. 
In this scene, the 20-band result shows weaker stripe artifacts. This observation is consistent with reduced sensitivity to some redundant or unstable bands, although no dedicated stripe-artifact metric is used here.

Following the region-based evaluation protocol used in prior work, three \(20\times40\) representative patches are selected from emission-dominant regions, including grass, foreground trees, and background forest. 
The grass and foreground-tree patches have corresponding LiDAR reference regions, as marked in Fig.~\ref{fig:fig61}(c). 
For the background forest region, reliable LiDAR reference is unavailable, so we mainly evaluate the concentration and stability of the estimated distance distribution.

Prior work reported bispectral, quadspectral, and hyperspectral ranging results on the same scene and evaluated local patches of the same size in the grass, foreground-tree, and background-forest regions~\cite{dorken_gallastegi_absorption-based_2025,gallastegi_ozone_2026}.
However, the exact coordinates of the selected patches were not specified.
Therefore, the reported values are used as literature reference values rather than strictly coordinate-aligned quantitative comparisons.
For a same-region comparison, we run the released hyperspectral ranging code and evaluate the reproduced result on the same regions selected in this paper.
Although the bispectral and quadspectral methods are computationally efficient, their reported ranging accuracy on this scene is lower than that of the hyperspectral formulation~\cite{dorken_gallastegi_absorption-based_2025,gallastegi_ozone_2026}.
Therefore, the reported and reproduced hyperspectral results are used as the main baseline comparisons.

Table~\ref{tab:tab2} summarizes the distance mean, standard deviation, and runtime of different methods on the three selected regions. 
For the grass and foreground-tree regions, the mean distances obtained by ADER full-band are closer to the LiDAR median than those obtained by the compared hyperspectral baselines. 
In particular, the LiDAR median distance of the foreground-tree region is \(57.48~\mathrm{m}\), while ADER full-band gives \(56.11~\mathrm{m}\pm10.08~\mathrm{m}\), showing a smaller bias than both the reported and reproduced hyperspectral results. 
Across the three selected regions, ADER full-band also gives the smallest standard deviation, indicating that its distance estimates are more concentrated. 
The 20-band result remains close to the full-band result while using far fewer spectral bands, which further supports the effectiveness of the proposed band selection strategy.
\textcolor{black}{
	However, a larger deviation is observed in the foreground-tree region, consistent with the weak temperature contrast and reduced spectral constraints discussed below.
}

\begin{table}[!t]
	\centering
	\caption{Quantitative comparison on two \(8\times8\) reflection-dominant patches.}
	\label{tab:tab3}
	\scriptsize
	\renewcommand{\arraystretch}{1.03}
	\setlength{\heavyrulewidth}{1.1pt}
	\setlength{\lightrulewidth}{0.55pt}
	
	\resizebox{\columnwidth}{!}{%
		\begin{tabular}{lcc}
			\toprule
			Method 
			& Front checkerboard
			& Rear checkerboard \\
			\midrule
			
			Decoupled only
			& \(180.51~\mathrm{m} \pm 8.68~\mathrm{m}\)
			& \(194.26~\mathrm{m} \pm 1.68~\mathrm{m}\) \\
			
			ADER full-band
			& \(\mathbf{30.41~\mathrm{m}} \pm 6.97~\mathrm{m}\)
			& \(\mathbf{44.87~\mathrm{m}} \pm 2.36~\mathrm{m}\) \\
			
			HS (reported)
			& \(28.86~\mathrm{m} \pm 5.38~\mathrm{m}\)
			& \(39.83~\mathrm{m} \pm 2.01~\mathrm{m}\) \\
			
			HS+TV (reported)
			& \(28.82~\mathrm{m} \pm 5.31~\mathrm{m}\)
			& \(39.84~\mathrm{m} \pm 1.78~\mathrm{m}\) \\
			
			LiDAR median
			& \(31.01~\mathrm{m}\)
			& \(46.51~\mathrm{m}\) \\
			
			\bottomrule
		\end{tabular}%
	}
	\vspace{-0.4em}
\end{table}	

In terms of computational efficiency, ADER full-band takes \(77.2~\mathrm{s}\) on this scene, while ADER 20-band takes \(56.23~\mathrm{s}\). 
Under the same testing condition, the released hyperspectral ranging code takes \(167.79~\mathrm{min}\). 
Therefore, the runtime of ADER full-band is reduced by approximately \(130\times\), and the 20-band setting further improves the efficiency. 
Overall, ADER achieves minute-level scene ranging and produces estimates that are closer to the LiDAR median and more stable in the evaluated regions.

It should be noted that the foreground-tree region remains a challenging case.
Although its mean distance is close to the LiDAR median, its standard deviation is still about \(10~\mathrm{m}\).
This dispersion is mainly related to the very low temperature contrast in this region, where the target--air temperature difference is below \(1~\mathrm{K}\), as well as the complex internal structure of vegetation.
For absorption-based ranging, when the target temperature approaches the air temperature, the observability of distance-related absorption structures decreases~\cite{dorken_gallastegi_absorption-based_2024}.
\textcolor{black}{
	Under such weak-distance-observability conditions, reducing the number of spectral bands from 256 to 20 further limits the spectral constraints available for inversion, consistent with the increased deviation and dispersion observed in the foreground-tree region.
}
ADER therefore does not eliminate the uncertainty in this weak-temperature-contrast region, but reduces the bias and improves the local stability compared with prior results.

To further analyze the local reliability of distance estimation in the real scene, we compute the local effective distance Fisher information based on the final ADER estimates, using the corresponding radiative models for emission-dominant and reflection-dominant pixels, respectively, and normalize it for visualization as a distance-observability indicator.
A larger effective Fisher information indicates stronger distance observability, as shown in Fig.~\ref{fig:fig61_observability}(a).
The low-Fisher-information regions are mainly concentrated in vegetation areas with weak target--air radiance contrast, including the lower part of the foreground trees and several regions in the background forest.
For emission-dominant pixels in these regions, distance sensitivity decreases
as the target self-emission \(\varepsilon(\lambda)B(\lambda,T)\) approaches
\(B(\lambda,T_{\mathrm{air}})\), weakening the radiance contrast that carries
distance-dependent modulation.
These low-observability regions also closely coincide with the locations exhibiting pronounced distance discontinuities in Fig.~\ref{fig:fig61}(a), showing consistent spatial distributions between anomalous distance estimates and insufficient local distance information.

Using the corresponding CRLB-derived local distance-uncertainty indicator, we further apply Tukey's interquartile-range-based rule for robust outlier identification~\cite{tukey_exploratory_1977}.
For the current scene, the resulting threshold is \(17.23~\mathrm{m}\), with \(6.55\%\) of the pixels flagged as low-information distance estimates.
The resulting full-scene map is shown in Fig.~\ref{fig:fig61_observability}(b), where the identified low-information pixels are flagged while the remaining distance estimates are left unchanged.
Further details on the Fisher-information calculation, CRLB-derived uncertainty indicator, and threshold definition are provided in Appendix~B-A of the supplementary material.
This flagging is used only for reliability diagnosis; to maintain the same evaluation support as the baseline methods, the quantitative comparison in Table~\ref{tab:tab2} continues to use the complete unfiltered regions.
\subsection{Reflection-Dominant Region Analysis}
\label{subsec:highly_reflective_region_analysis}

To further analyze distance estimation in reflection-dominant regions, this subsection conducts the evaluation from two aspects. 
First, we use the same regions as those defined in the released code of the prior method to evaluate the effect of reflection-compensated refinement on local ranging accuracy.
Second, we examine larger reflective-board regions to assess the effect of the B-spline emissivity representation on spatial consistency.

Distance estimation for reflection-dominant pixels requires the complete radiative model that includes the reflected downwelling-radiance term. 
Specifically, we use the Adam optimizer together with an externally simulated downwelling-radiance dictionary to refine the detected reflection-dominant pixels, and jointly estimate the target temperature, distance, emissivity control coefficients, and downwelling-radiance coefficients. 
This refinement follows the downwelling-radiance dictionary modeling used in prior LWIR absorption-based ranging work~\cite{gallastegi_ozone_2026}.

Fig.~\ref{fig:fig62} shows the correction effect of reflection-compensated refinement on the reflective-board regions. 
When only the decoupled estimation is used, the reflective-board regions suffer from clear distance overestimation. 
After reflection-compensated refinement, the overestimation is reduced and the distance distribution on each reflective board becomes more spatially uniform.

For comparability with prior work~\cite{gallastegi_ozone_2026}, we use two \(8\times8\) pixel patches whose locations and sizes are exactly the same as the evaluation regions specified in the released code of the prior method. 
Table~\ref{tab:tab3} reports the mean and standard deviation of the estimated distances in these two regions. 
Using only the decoupled estimate leads to severe distance overestimation in the reflection-dominant regions, indicating that the simplified model without reflection cannot explain the influence of downwelling radiance in these pixels. 
After reflection-compensated refinement is introduced, ADER full-band obtains distance estimates closer to the LiDAR median in the selected regions. 
By contrast, although HS+TV reduces the local standard deviation, its mean distance still shows evident underestimation. 
Overall, ADER full-band obtains smaller mean bias on both the front and rear checkerboards while maintaining reasonable local stability.

The above evaluation focuses on local accuracy. 
To assess spatial consistency across reflective boards, we further select larger rectangular regions covering different material subregions. 
Since prior work did not report statistics on these regions, we use the reproduced result as the baseline. 
We also construct ADER-per-band by replacing the B-spline emissivity representation with independent per-band emissivity variables while keeping the same refinement procedure.

\begin{figure}[!t]
	\centering
	\setlength{\abovecaptionskip}{2pt}
	\setlength{\belowcaptionskip}{0pt}
	
	\begin{minipage}[t]{0.290\linewidth}
		\centering
		\includegraphics[width=\linewidth]{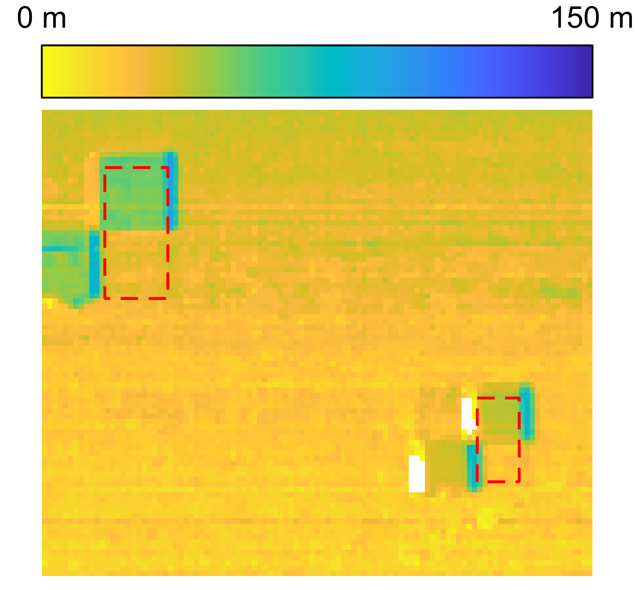}
		\vspace{-1.5mm}
		\centerline{\scriptsize (a) Reproduced HS}
		\label{fig:fig63a}
	\end{minipage}%
	\hspace{0.012\linewidth}%
	\begin{minipage}[t]{0.290\linewidth}
		\centering
		\includegraphics[width=\linewidth]{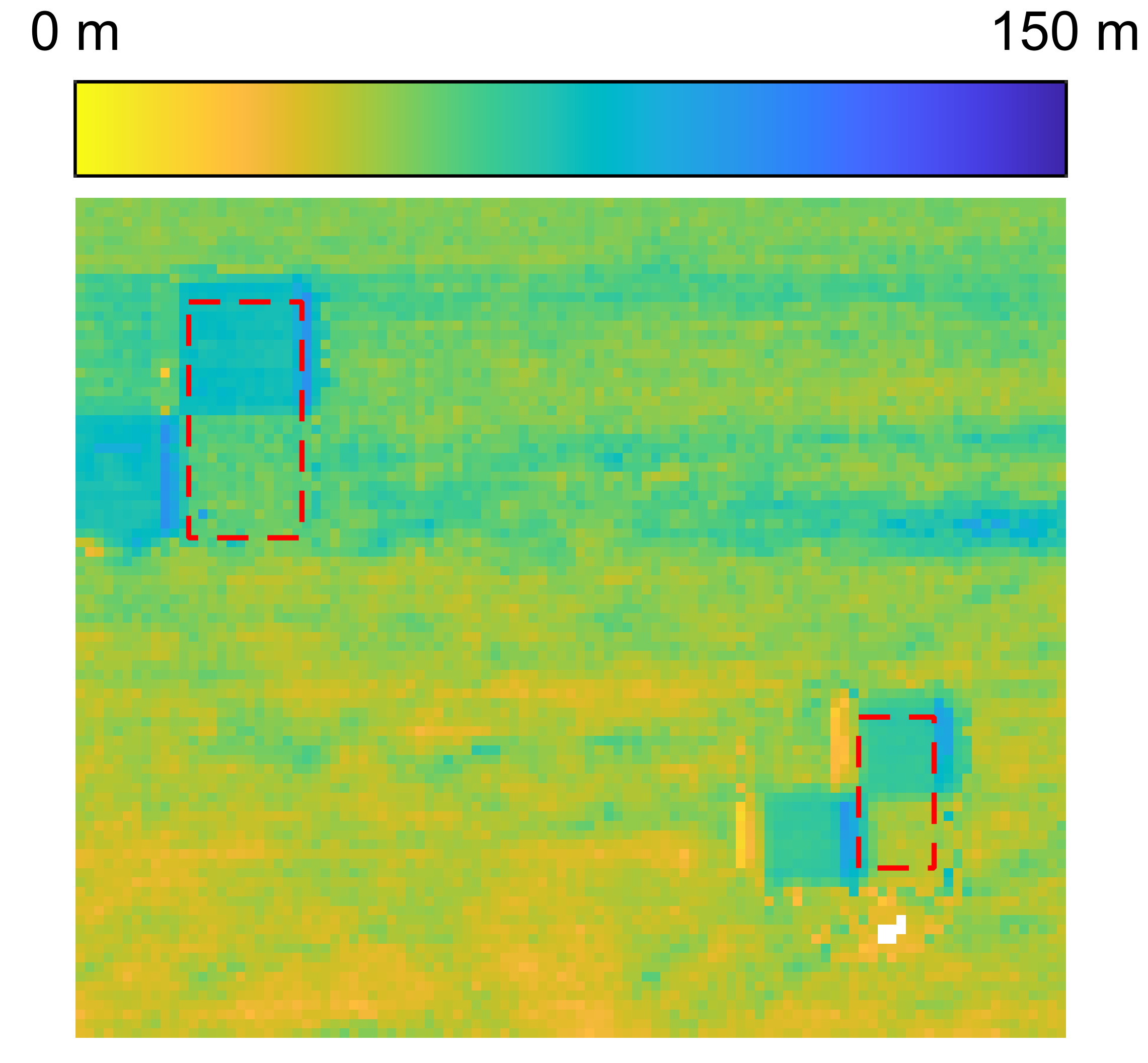}
		\vspace{-1.5mm}
		\centerline{\scriptsize (b) ADER-per-band}
		\label{fig:fig63b}
	\end{minipage}%
	\hspace{0.012\linewidth}%
	\begin{minipage}[t]{0.290\linewidth}
		\centering
		\includegraphics[width=\linewidth]{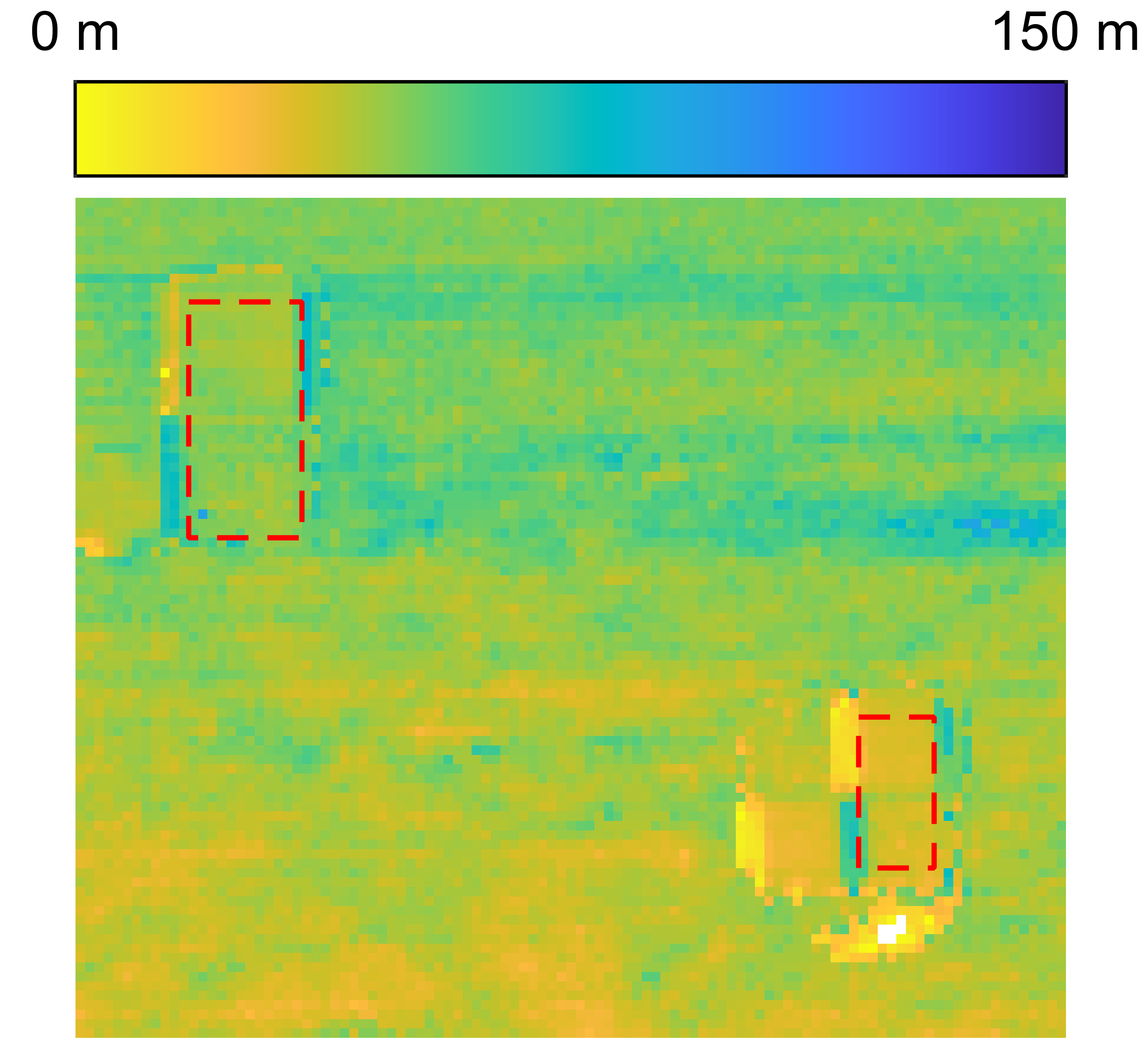}
		\vspace{-1.5mm}
		\centerline{\scriptsize (c) ADER full-band}
		\label{fig:fig63c}
	\end{minipage}
	
	\vspace{0.8mm}
	
	\begin{minipage}[t]{0.475\linewidth}
		\centering
		\includegraphics[width=\linewidth]{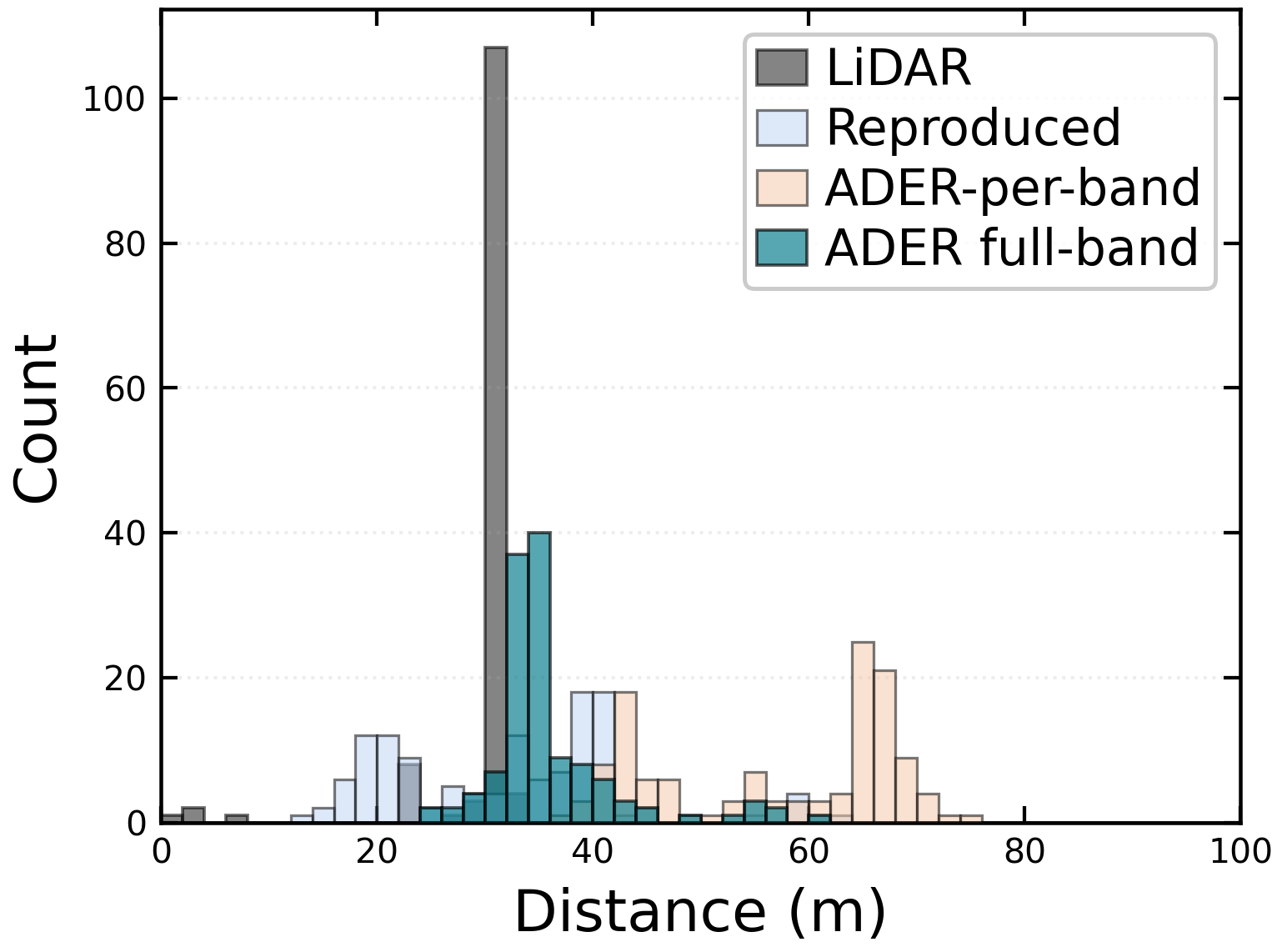}
		\vspace{-1.5mm}
		\centerline{\scriptsize (d) Front checkerboard}
		\label{fig:fig63d}
	\end{minipage}%
	\hspace{0.015\linewidth}%
	\begin{minipage}[t]{0.475\linewidth}
		\centering
		\includegraphics[width=\linewidth]{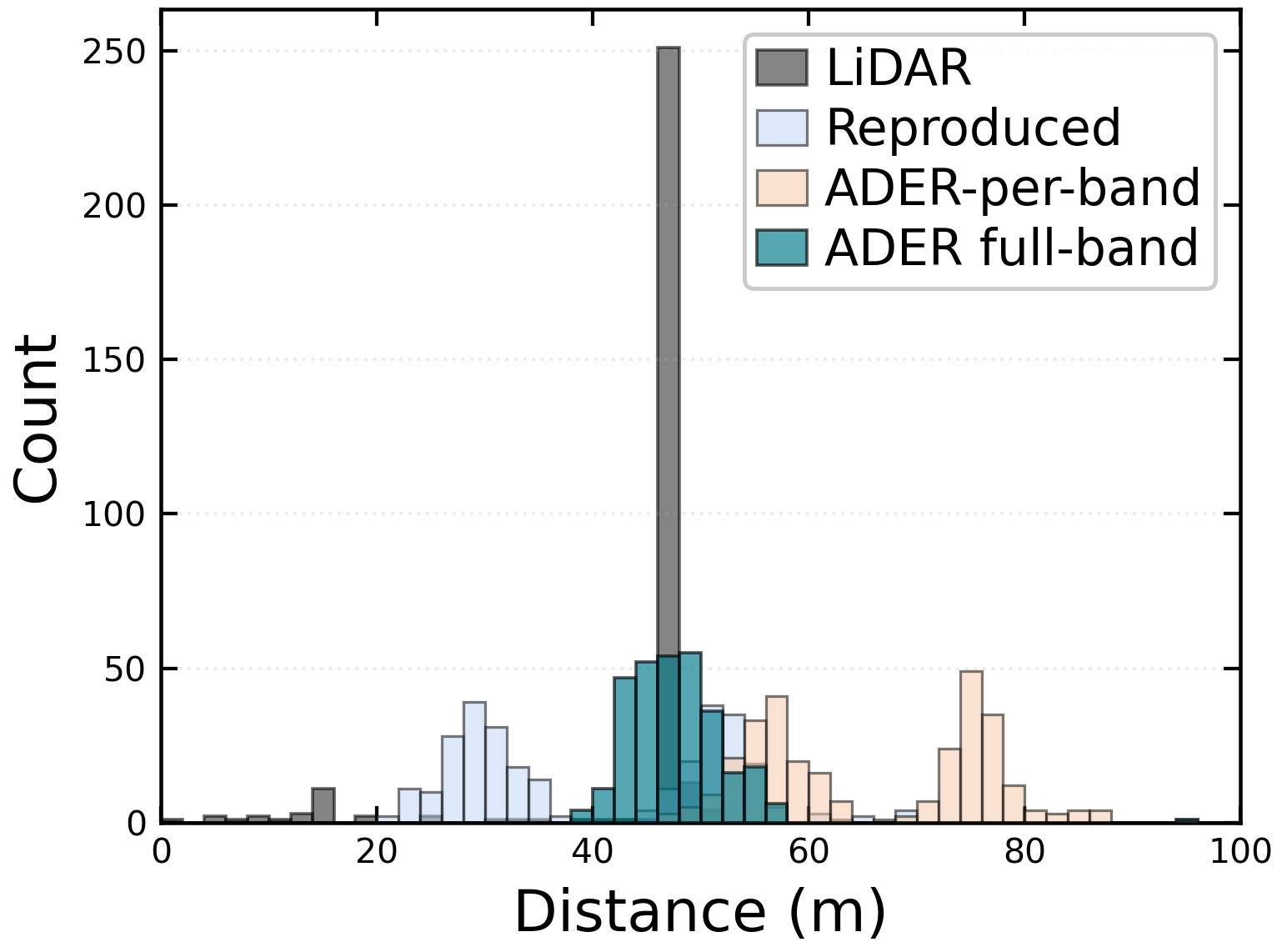}
		\vspace{-1.5mm}
		\centerline{\scriptsize (e) Rear checkerboard}
		\label{fig:fig63e}
	\end{minipage}
	
	\vspace{-0.4em}
	\caption{Reflective-board ranging and histogram comparison.
		(a) Reproduced HS baseline.
		(b) ADER-per-band.
		(c) ADER full-band.
		(d),(e) Distance histograms of the front and rear checkerboards.
		Red boxes mark the patches used in Table~\ref{tab:tab4}.}
	\label{fig:fig63}
	\vspace{-0.8em}
\end{figure}

Fig.~\ref{fig:fig63} compares the HS baseline (reproduced), the ADER-per-band ablation variant, and ADER full-band on two reflective-board patches, together with the corresponding distance histograms.
Both the HS baseline and ADER-per-band show clear within-board distance discontinuities, indicating unstable estimation in reflection-dominant regions.
For ADER-per-band, the independent per-band emissivity variables introduce excessive spectral degrees of freedom, making distance inversion more susceptible to the coupled effects of temperature, emissivity, and downwelling radiance.
\textcolor{black}{
	Although the reproduced HS baseline also estimates emissivity independently at each band, it incorporates spectral smoothness regularization, whereas ADER-per-band retains ADER's reflection-compensated refinement without the B-spline constraint.
	These differences in regularization and inversion strategy can contribute to the larger mean-distance deviations of ADER-per-band relative to the HS baseline.
}
In contrast, ADER full-band uses the B-spline emissivity representation to constrain emissivity to a smooth low-dimensional space, thereby reducing this ambiguity and producing more spatially consistent distances and more concentrated histograms.
Table~\ref{tab:tab4} further confirms this improvement through the lower standard deviations on both reflective-board patches.

\begin{table}[!t]
	\centering
	\caption{Distance distribution evaluation on selected reflection-dominant patches.}
	\label{tab:tab4}
	\scriptsize
	\renewcommand{\arraystretch}{1.03}
	\setlength{\heavyrulewidth}{1.1pt}
	\setlength{\lightrulewidth}{0.55pt}
	
	\resizebox{\columnwidth}{!}{%
		\begin{tabular}{lcc}
			\toprule
			Method 
			& Front checkerboard
			& Rear checkerboard \\
			\midrule
			
			HS baseline (reproduced)
			& \(32.15~\mathrm{m} \pm 10.83~\mathrm{m}\)
			& \(40.51~\mathrm{m} \pm 12.22~\mathrm{m}\) \\
			
			ADER-per-band
			& \(57.14~\mathrm{m} \pm 10.86~\mathrm{m}\)
			& \(65.87~\mathrm{m} \pm 10.69~\mathrm{m}\) \\
			
			ADER full-band
			& \(36.04~\mathrm{m} \pm \mathbf{6.12~\mathrm{m}}\)
			& \(47.74~\mathrm{m} \pm \mathbf{4.70~\mathrm{m}}\) \\
			
			LiDAR median
			& \(31.01~\mathrm{m}\)
			& \(46.51~\mathrm{m}\) \\
			
			\bottomrule
		\end{tabular}%
	}
	\vspace{-0.4em}
\end{table}

\subsection{Reduced-Band Ranging and Cross-Scene Validation}
\label{subsec:reduced_band_ranging_analysis}
\begin{figure}[!t]
	\centering
	\setlength{\abovecaptionskip}{2pt}
	\setlength{\belowcaptionskip}{0pt}
	
	\begin{minipage}[t]{0.48\columnwidth}
		\centering
		\makebox[\linewidth][c]{\scriptsize ADER full-band}
		\vspace{0.8mm}
		\includegraphics[width=\linewidth]{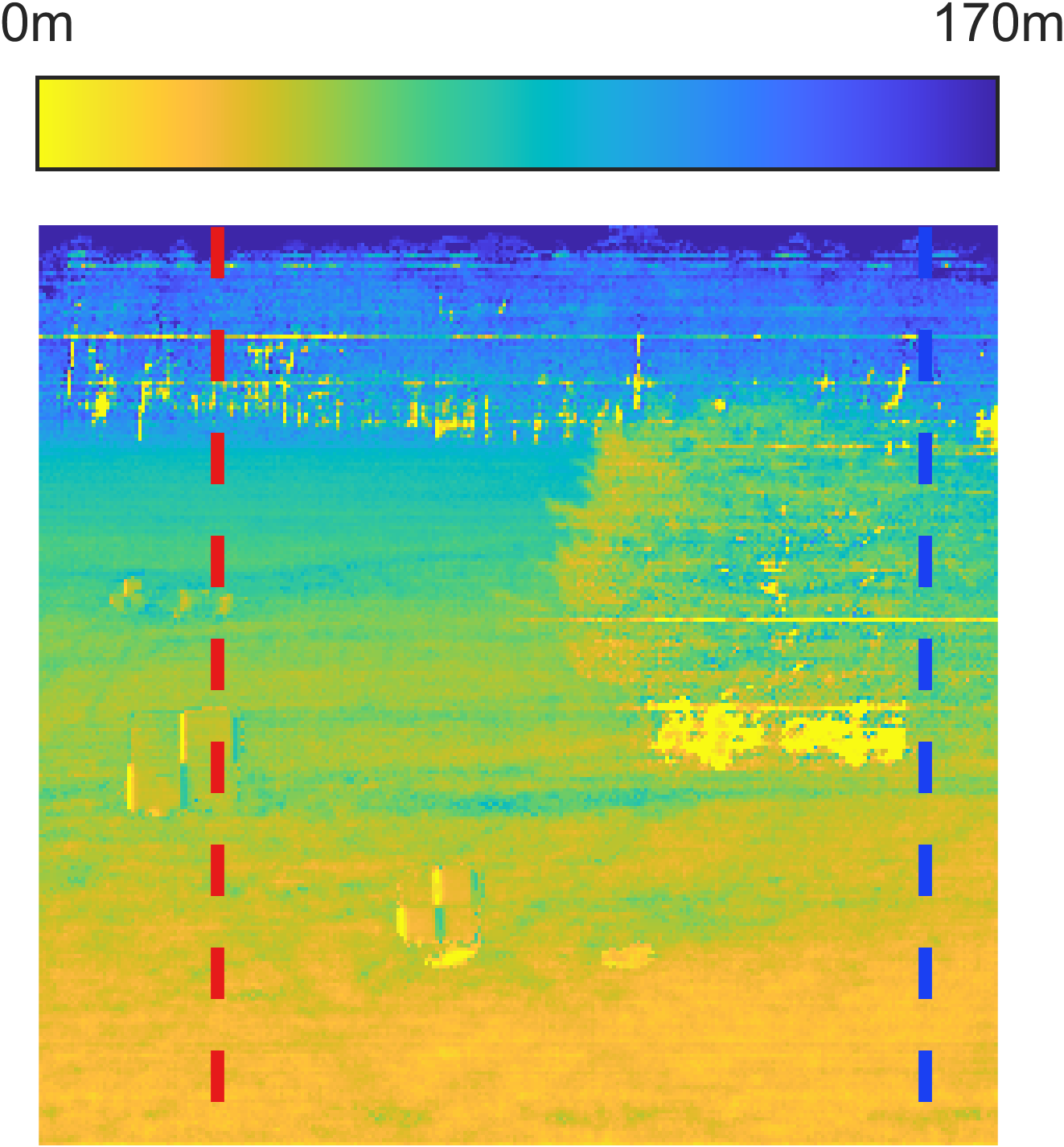}
	\end{minipage}
	\hfill
	\begin{minipage}[t]{0.48\columnwidth}
		\centering
		\makebox[\linewidth][c]{\scriptsize ADER 20-band}
		\vspace{0.8mm}
		\includegraphics[width=\linewidth]{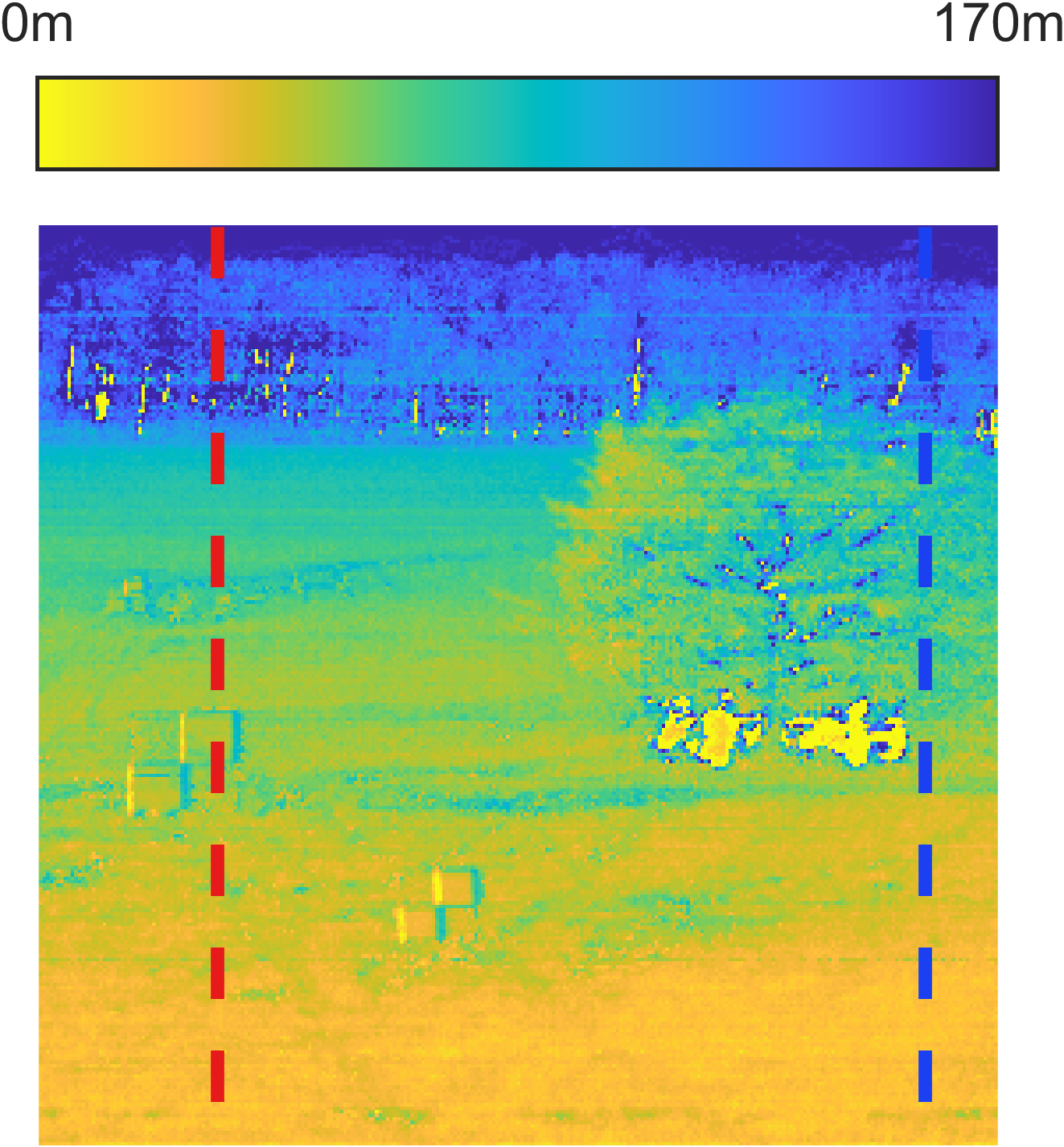}
	\end{minipage}
	
	\vspace{0.8mm}
	\makebox[\columnwidth][c]{\scriptsize (a) Profile locations}
	
	\vspace{1.0mm}
	
	\includegraphics[width=0.96\columnwidth]{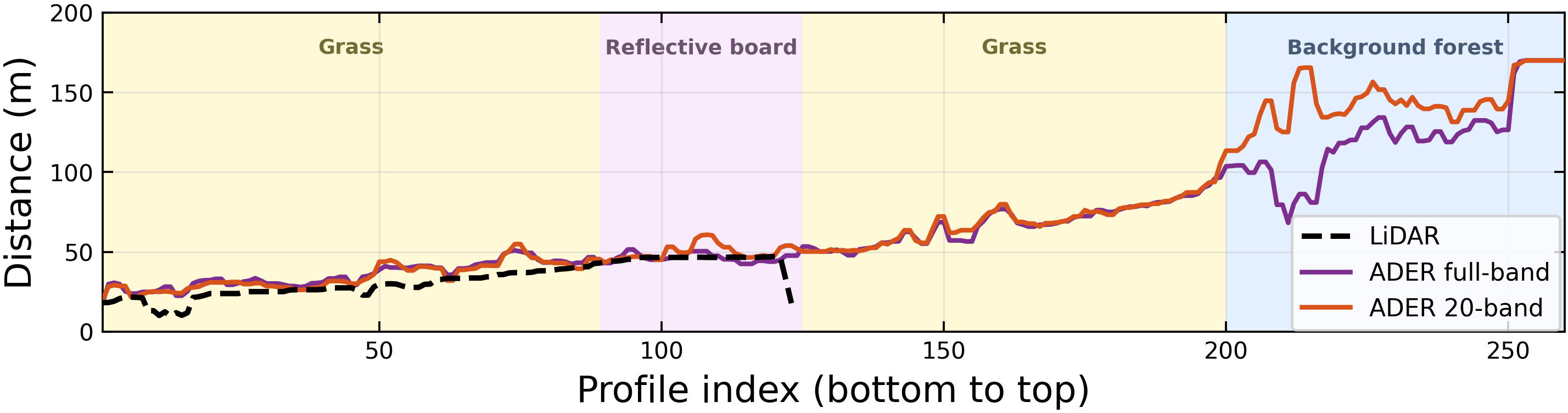}
	\vspace{-0.4mm}
	\makebox[\columnwidth][c]{\scriptsize (b) Red profile}
	
	\vspace{0.9mm}
	
	\includegraphics[width=0.96\columnwidth]{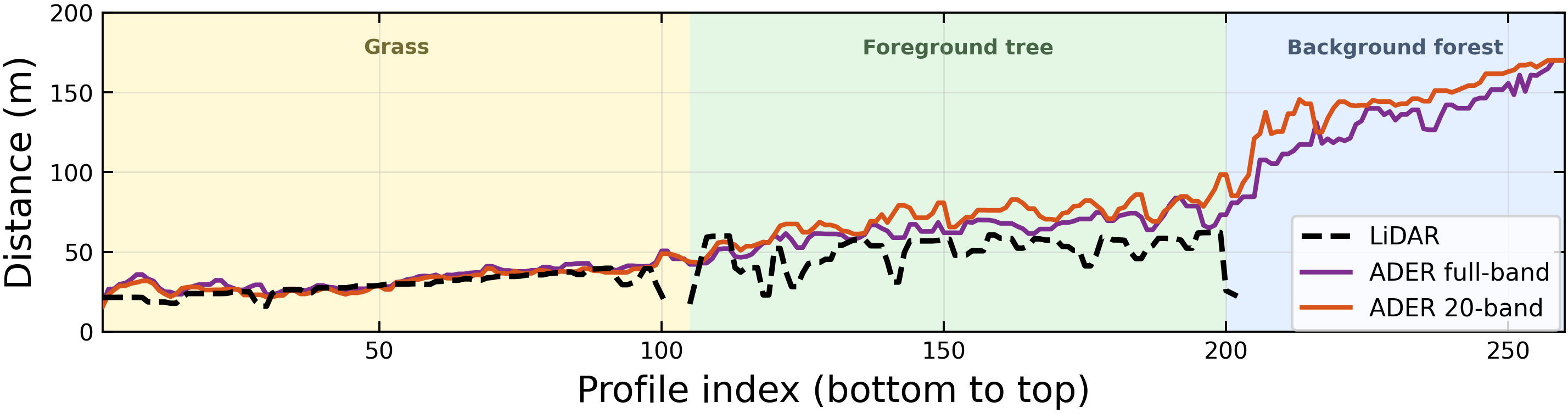}
	\vspace{-0.4mm}
	\makebox[\columnwidth][c]{\scriptsize (c) Blue profile}
	
	\vspace{1mm}
	\caption{Profile comparison between ADER full-band and ADER 20-band results.
		(a) Profile locations.
		(b),(c) Red and blue profiles, where the red profile passes through near grass, reflective boards, and background forest, while the blue profile mainly passes through grass, foreground trees, and background forest.
		The dashed curve denotes LiDAR, and the solid curves denote ADER results.
		Profile indices increase from bottom to top.}
	\label{fig:fig64}
	\vspace{-0.6em}
\end{figure}

Based on the greedy band selection strategy proposed in Section~\ref{sec:band_selection}, we further evaluate reduced-band ranging on real scenes. 
The representative-region evaluation in Table~\ref{tab:tab2} shows that ADER 20-band achieves ranging results close to ADER full-band using 256 bands, while substantially reducing the number of spectral bands. 
To further analyze the spatial consistency of ADER 20-band, this subsection first compares the distance variations of full-band and 20-band results along selected profile lines, and then evaluates the applicability of the selected bands on multiple real scenes.

Fig.~\ref{fig:fig64}(a) shows the ranging maps used for profile analysis, where the left image is the ADER full-band result and the right image is the ADER 20-band result. 
The red and blue dashed lines denote two selected vertical profiles that cover different scene structures. 
The red profile passes through near grass, reflective boards, and background forest, and is used to analyze distance variations under a mixture of emission-dominant and reflection-dominant pixels, as shown in Fig.~\ref{fig:fig64}(b). 
The blue profile mainly passes through grass, foreground trees, and background forest, and is used to evaluate ranging stability under weak temperature contrast and the separability between foreground trees and distant background forest, as shown in Fig.~\ref{fig:fig64}(c).

As shown in Fig.~\ref{fig:fig64}, ADER full-band and ADER 20-band both follow distance variation trends close to the LiDAR reference along the main regions of the two profiles. 
In particular, within the near-range region below \(100~\mathrm{m}\), the results before and after band selection are largely consistent, indicating that the selected bands preserve the main distance-sensitive information. 
In challenging regions, including weak-temperature-contrast foreground trees, background forest, and reflective boards, ADER full-band produces smoother curves overall, while ADER 20-band still maintains a similar spatial trend without evident structural deviation. 
It should be noted that the LiDAR range map and the LWIR hyperspectral image are only approximately registered, so local differences between the profile curves may also be affected by registration mismatch. 
Overall, the profile analysis shows that ADER 20-band preserves spatial consistency close to the full-band result in the real scene, supporting the effectiveness of the proposed task-driven band selection strategy.

\begin{figure*}[!t]
	\color{black}
	\centering
	\setlength{\abovecaptionskip}{4pt}
	\setlength{\belowcaptionskip}{0pt}
	
	\includegraphics[width=0.78\textwidth]{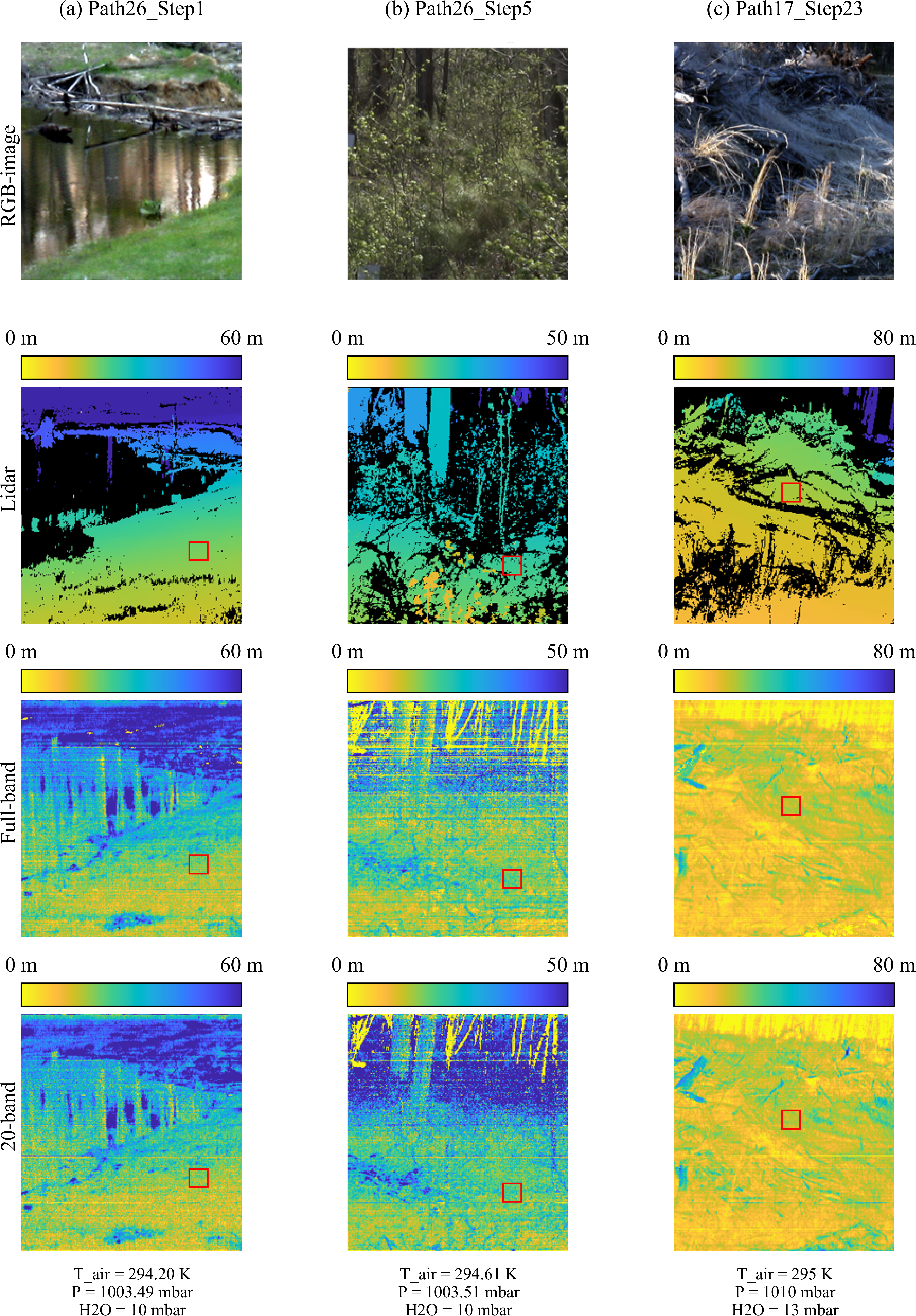}
	
	\vspace{1mm}
	\caption{Cross-scene ranging validation on additional IH dataset scenes.
		The rows show the visible reference image, approximately registered LiDAR reference, ADER full-band result, and ADER 20-band result, respectively.
		The red boxes denote the \(20\times20\) patches used for Table~\ref{tab:tab5}.
		The MODTRAN atmospheric parameters are marked below each scene.}
	\label{fig:fig65}
	\vspace{-0.6em}
\end{figure*}

Fig.~\ref{fig:fig65} shows cross-scene results on three additional IH scenes, with the MODTRAN atmospheric parameters marked below each scene. 
ADER 20-band preserves the main spatial structures of ADER full-band and shows trends similar to the LiDAR reference. 
For quantitative evaluation, Table IV reports the mean and standard deviation of ADER full-band and ADER 20-band on selected $20\times20$ patches, together with the LiDAR median. The 20-band result maintains an error level similar to the full-band result, indicating that the selected bands preserve the main distance information across scenes. \textcolor{black}{Further enlarged-region evaluations, together with additional cross-scene results and a discussion of practical limitations, are provided in Appendix G-B of the supplementary material.}

\begin{table}[!t]
	\centering
	\color{black}
	\caption{Distance statistics on \(20\times20\) selected patches from additional scenes.
			Bold mean values denote the ADER result closest to the LiDAR median in each scene.}
	\label{tab:tab5}
	\scriptsize
	\renewcommand{\arraystretch}{1.08}
	\setlength{\heavyrulewidth}{1.0pt}
	\setlength{\lightrulewidth}{0.5pt}
	
	{\color{black}
		\resizebox{\columnwidth}{!}{%
			\begin{tabular}{lccc}
				\toprule
				Method
				& Path26\_Step1
				& Path26\_Step5
				& Path17\_Step23 \\
				\midrule
				
				ADER full-band
				& \(\mathbf{20.97}~\mathrm{m} \pm 3.76~\mathrm{m}\)
				& \(18.41~\mathrm{m} \pm 3.53~\mathrm{m}\)
				& \(21.27~\mathrm{m} \pm 2.82~\mathrm{m}\) \\
				
				ADER 20-band
				& \(19.85~\mathrm{m} \pm 4.32~\mathrm{m}\)
				& \(\mathbf{19.17}~\mathrm{m} \pm 3.81~\mathrm{m}\)
				& \(\mathbf{21.33}~\mathrm{m} \pm 3.18~\mathrm{m}\) \\
				
				LiDAR median
				& \(20.56~\mathrm{m}\)
				& \(20.98~\mathrm{m}\)
				& \(23.20~\mathrm{m}\) \\
				
				\bottomrule
			\end{tabular}%
	}}
	\vspace{-0.4em}
\end{table}

\section{Conclusion}
\label{sec:conclusion}

This paper presents ADER for LWIR hyperspectral passive ranging in natural scenes, where distance is nonlinearly coupled with temperature, material emissivity, path radiance, and reflected downwelling radiance. 
ADER uses atmospheric absorption structures to decouple distance estimation, rather than relying solely on full-band joint inversion. 
By representing emissivity with B-spline control points, classifying pixels according to downwelling-radiance cues, and selecting distance-sensitive bands using multi-scene effective Fisher information for the distance parameter, ADER reduces the dimensionality and spectral redundancy of the ranging problem while preserving the absorption information needed for distance estimation.

Experiments on real LWIR hyperspectral scenes show that ADER recovers spatial distance structures that are consistent with the approximately registered LiDAR reference. 
On the main scene, ADER provides more accurate local distance statistics than the public full-band hyperspectral ranging implementation and reduces the runtime from \(167.79~\mathrm{min}\) to \(77.2~\mathrm{s}\), corresponding to an approximately \(130\times\) speedup. 
The analysis of reflection-dominant regions further shows that reflection-compensated refinement reduces distance overestimation, while the B-spline emissivity representation improves the spatial consistency of the estimated distances on reflection-dominant regions. 
The reduced-band experiments indicate that ADER with 20 selected bands can maintain distance structures close to the full-band result in the evaluated scenes, supporting the use of task-driven band selection for LWIR hyperspectral passive ranging.

	Despite these results, the practical reliability of ADER remains fundamentally limited by the observability of distance information in absorption-based passive ranging. 
	Specifically, the foreground-tree region in the present experiment, where the target--air temperature difference is below \(1~\mathrm{K}\), represents a challenging low-observability case. 
	More generally, consistent with the full-scene distance-observability analysis in Section~VI-B, when the target--air radiance contrast is weak or when atmospheric absorption along the propagation path is insufficient to produce distinct distance-dependent spectral modulation, the observability of distance information decreases and the spectral differences among candidate distances become less pronounced. 
	Under such conditions, distance estimation becomes more susceptible to sensor noise, complex scene structure, and model mismatch, thereby reducing the stability and reliability of the ranging results.

	In addition, the current ADER implementation still relies on scene-level atmospheric priors and MODTRAN-simulated atmospheric transmittance. 
	When the adopted atmospheric parameters deviate from the actual atmospheric state during imaging, modeling errors in transmittance and path radiance may propagate into the distance estimation process, thereby affecting ranging accuracy and stability and limiting cross-scene applicability. 
	Future work will investigate adaptive atmospheric parameter estimation and scene-adaptive atmospheric correction, improve ranging reliability under low-observability conditions, and conduct broader cross-scene validation and system-level experiments using calibrated LWIR hyperspectral sensors.

\ifCLASSOPTIONcaptionsoff
\newpage
\fi

\bibliographystyle{IEEEtran}
\bibliography{refs}

\end{document}